\documentclass[lettersize,journal]{IEEEtran}
\usepackage{amsmath,amsfonts}
\usepackage{algorithmic}
\usepackage{algorithm}
\usepackage{array}
\usepackage[caption=false,font=normalsize,labelfont=sf,textfont=sf]{subfig}
\usepackage{textcomp}
\usepackage{stfloats}
\usepackage{url}
\usepackage{verbatim}
\usepackage{graphicx}
\usepackage{tabularx}
\usepackage{multirow}
\usepackage{booktabs}
\usepackage{xcolor}

\usepackage{booktabs}
\usepackage{colortbl}
\usepackage{graphicx} % for \resizebox if needed
\usepackage{pifont} % Add this to the preamble for checkmarks and crosses
\newcommand{\cmark}{\textcolor{green!60!black}{\ding{52}}}  % green check
\newcommand{\xmark}{\textcolor{red!75!black}{\ding{55}}}    % red cross
\usepackage{cite}
\usepackage[colorlinks=true,
            linkcolor=green,       % for \ref and \pageref
            citecolor=green,       % for \cite
            urlcolor=blue]         % for \url and \href
            {hyperref}
            
\begin{document}

\title{Unsupervised Anomaly Detection for Autonomous Robots via Mahalanobis SVDD with Audio-IMU Fusion}

\author{Yizhuo Yang\textsuperscript{*}, Jiulin Zhao\textsuperscript{*}, Xinhang Xu, Kun Cao, Shenghai Yuan,~\IEEEmembership{Member,~IEEE} and Lihua Xie,~\IEEEmembership{Fellow,~IEEE}
        % <-this % stops a space
        
\thanks{* Joint first authors with equal contribution}% <-this % stops a space
\thanks{The authors are with the School of Electrical and Electronic Engineering, Nanyang Technological University, 50 Nanyang Avenue, Singapore 639798.  (email: yang0670@e.ntu.edu.sg, jzhao030@e.ntu.edu.sg, xinhang.xu@ntu.edu.sg, kun001@e.ntu.edu.sg, shyuan@ntu.edu.sg,  elhxie@ntu.edu.sg).}}



\maketitle

\begin{abstract}
Reliable anomaly detection is essential for ensuring the safety of autonomous robots, particularly when conventional detection systems based on vision or LiDAR become unreliable in adverse or unpredictable conditions. In such scenarios, alternative sensing modalities are needed to provide timely and robust feedback. To this end, we explore the use of audio and inertial measurement unit (IMU) sensors to detect underlying anomalies in autonomous mobile robots, such as collisions and internal mechanical faults.
Furthermore, to address the challenge of limited labeled anomaly data, we propose an unsupervised anomaly detection framework based on Mahalanobis Support Vector Data Description (M-SVDD). In contrast to conventional SVDD methods that rely on Euclidean distance and assume isotropic feature distributions, our approach employs the Mahalanobis distance to adaptively scale feature dimensions and capture inter-feature correlations, enabling more expressive decision boundaries. In addition, a reconstruction-based auxiliary branch is introduced to preserve feature diversity and prevent representation collapse, further enhancing the robustness of anomaly detection.
Extensive experiments on a collected mobile robot dataset and four public datasets demonstrate the effectiveness of the proposed method, as shown in the video \url{https://youtu.be/yh1tn6DDD4A}.
Code and dataset are available at \url{https://github.com/jamesyang7/M-SVDD}.
\end{abstract}

\begin{IEEEkeywords}
Anomaly detection, audio-IMU fusion, unsupervised learning, support vector data description
\end{IEEEkeywords}

\section{Introduction}
% \IEEEPARstart{A}{utonomous} mobile robots are increasingly deployed in diverse real-world environments, including industrial automation, logistic distribution, and human-robot collaborative tasks. In these complex and dynamic environments, robots inevitably encounter unexpected incidents, such as accidental collisions or internal component malfunctions, which pose severe threats to operational safety and reliability. Consequently, precise anomaly detection is essential for maintaining safe operations and preventing potential hazards.
\IEEEPARstart{A}{nomaly} detection is essential for ensuring the safety and reliability of many safety-critical systems, including industrial automation \cite{liu2024deep}, aerospace \cite{kraljevski2021machine} and robotic systems \cite{park2018multimodal,yoo2021multimodal}. In the context of autonomous mobile robots, the ability to detect anomalies, such as mechanical faults or unexpected collisions, is important for maintaining safe operations and preventing potential hazards. With the increasing deployment of robots in dynamic and unstructured environments, robust anomaly detection has become an indispensable component of autonomous operation.

Recent advances in anomaly detection for robotic systems have predominantly relied on vision sensors (RGB cameras) \cite{wellhausen2020safe}  and LiDAR \cite{ji2022proactive}. These methods achieve impressive performance in controlled settings but suffer significant degradation under adverse environmental conditions \cite{sezgin2023safe}, including changes in illumination, extreme weathers, occlusions, and sensor degradation. Moreover, visual and LiDAR-based methods typically focus on semantic or spatial anomalies, which may not directly capture internal system failures such as mechanical issues or vibration abnormalities. In addition, many of these methods depend on costly sensing equipment and computationally intensive data processing pipelines, limiting their practicality for deployment on platforms with constrained resources.
% Once these predictive methods fail to detect the underlying anomalies, a reliable anomaly detection mechanism is essential to monitor the robot's physical state and ensure appropriate responses. 
% While vision \cite{wellhausen2020safe} and LiDAR-based \cite{ji2022proactive} methods have been widely explored for anomaly detection, these predictive approaches exhibit notable limitations. They are often sensitive to environmental disturbances \cite{sezgin2023safe}, such as changes in illumination, adverse weather conditions, occlusions, and sensor degradation, which significantly reduce their reliability in unpredictable or degraded scenarios. 
% In addition, many of these methods depend on costly sensing equipment and computationally intensive data processing pipelines, limiting their practicality for deployment on platforms with constrained resources.
\begin{figure}[t]
    \centering
    \includegraphics[width=1\linewidth]{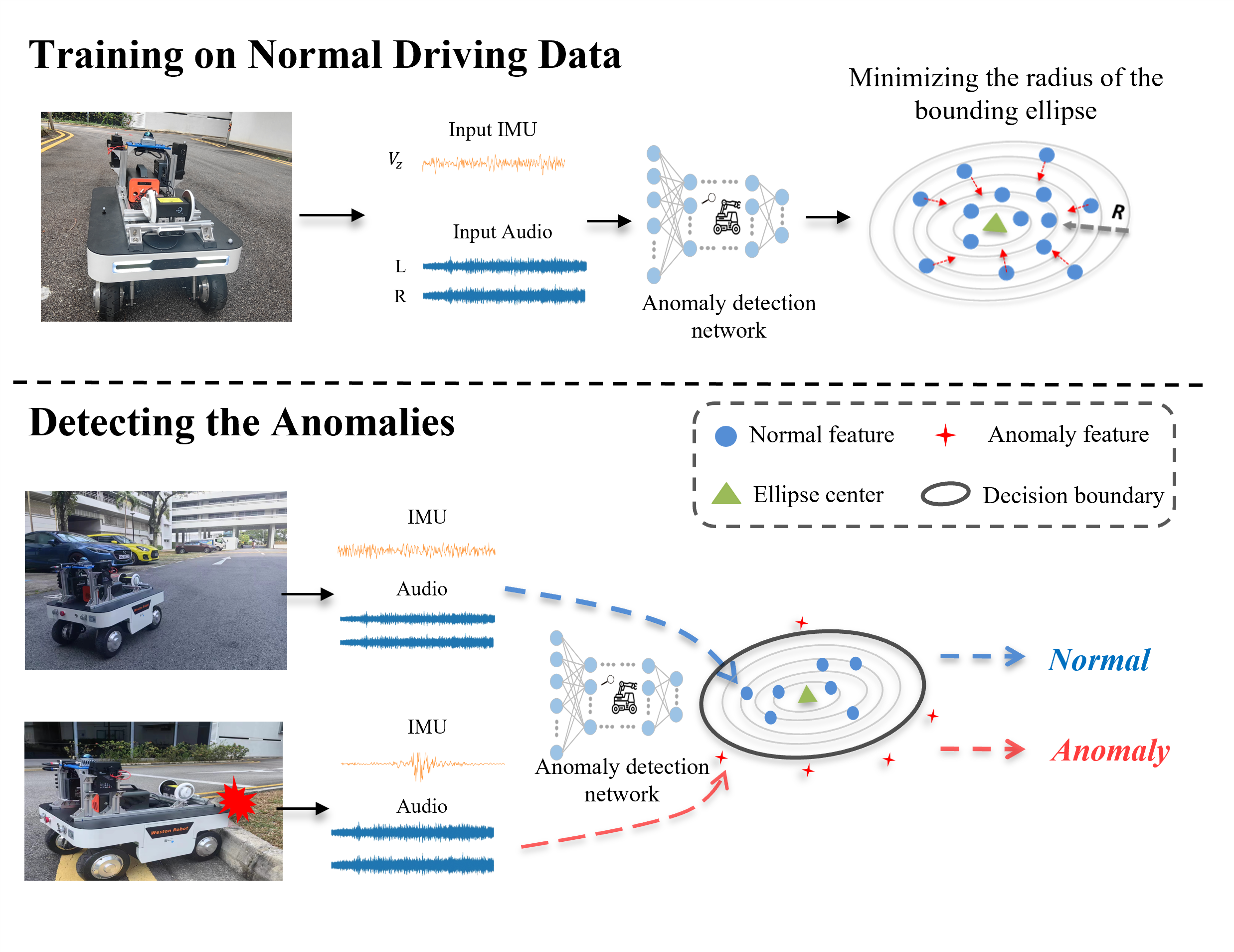}
    \caption{Our proposed method integrates audio and IMU data for anomaly detection in autonomous robots. The network is trained exclusively on normal operational data and effectively identifies anomalies during inference.}
    % \vspace{-2mm}
    \label{overview}
\end{figure}

Alternatively, audio and inertial measurement unit (IMU) sensors represent valuable, underexplored modalities for anomaly detection within robotic systems. Acoustic sensors naturally capture distinctive sound signatures generated by abnormal events, such as collisions, frictional contact, or internal mechanical faults. IMU sensors similarly detect anomalous vibrations and accelerations arising from collisions or mechanical failures, thus offering complementary information to audio data. Furthermore, both audio and IMU sensors are energy-efficient and cost-effective, making them particularly suitable for deployment in robotic systems.

Another fundamental challenge in anomaly detection for autonomous robots is the limited availability of anomaly data. Most current anomaly detection approaches \cite{li2024novel,zhang2023fault} rely on supervised learning techniques, requiring large-scale annotated datasets containing sufficient examples of diverse anomalies. However, collecting a large amount of labeled anomaly data in robotic systems is highly impractical. Robots are predominantly operated under safe and normal conditions, where abnormal scenarios are rarely encountered. Moreover, deliberately inducing faults or unsafe events for data collection is often infeasible, costly, or unethical. 
To address the scarcity of labeled data, some studies have explored semi-supervised \cite{10900541} and few-shot learning \cite{wang2024few} techniques, which have shown promising performance under limited data settings. Nevertheless, anomalies encountered in real-world environments are highly unpredictable and diverse, and these methods often struggle to generalize to previously unseen fault types, especially when the anomaly characteristics differ significantly from the training distribution. 
Thus, it is crucial to develop an unsupervised anomaly detection method that eliminates the need for labeled anomaly data while maintaining strong generalization capabilities to detect different types of anomalies.

Motivated by these challenges, this paper introduces an unsupervised framework that effectively fuses audio and IMU sensor data for mobile robot anomaly detection as shown in Fig. \ref{overview}. Specifically, we develop a Mahalanobis distance \cite{rippel2021modeling}-based Deep Support Vector Data Description (M-SVDD) model that learns to encapsulate normal driving data within an ellipsoidal boundary. During inference, anomalies are identified based on their Mahalanobis distance from the center of the learned distribution. To further improve representation quality and prevent feature collapse within the latent space, a reconstruction branch is incorporated as an auxiliary task to improve the detection robustness and accuracy.

The contribution of this paper can be summarized as follows:
\begin{itemize}
    \item We propose a novel unsupervised anomaly detection network to identify anomalies in autonomous robots. This network integrates audio and IMU data to effectively detect anomalies such as collisions and mechanical faults using only normal data for training.
    
    \item We propose a Mahalanobis SVDD model that leverages Mahalanobis distance and robust covariance estimation to learn an adaptive ellipsoidal boundary, while a reconstruction branch is incorporated to enhance feature representations and improve detection performance.
    
    \item We construct a new multimodal anomaly detection dataset comprising audio, IMU, LiDAR, and image data collected under both normal and fault conditions. We evaluate the proposed Mahalanobis SVDD model on this developed dataset as well as four public datasets, demonstrating superior performance over existing baselines.
    
\end{itemize}

\section{Related Work}
\subsection{Unsupervised Anomaly Detection}
Unsupervised anomaly detection aims to find the outliers of the given signals and identify the potential faults of the system without using any prior knowledge of the anomaly data. It has gained increasing attention in many fields, such as industrial manufacturing \cite{liu2024deep,brockmann2023voraus} and state monitoring of machine health \cite{peng2025reconstruction}. For example, Carrat\`u et al. \cite{carratu2023novel} develop an unsupervised anomaly detection framework for electrical systems by using K-means clustering on short-time Fourier transform (STFT)-transformed electrical signals. Guo et al. \cite{8603356} propose a One-class support vector machine (OCSVM)-based anomaly detection method for industrial applications by optimizing the decision boundary using a clustering-based local outlier detection method.
With the rapid development of deep learning technologies, many unsupervised methods based on deep neural networks have been proposed. For instance, Xu et al. \cite{xu2024calibrated} propose calibrated one-class classification for time series data anomaly detection by incorporating uncertainty-based calibration and synthetic anomaly generation to enhance the precision of the learned normality boundary. Tian et al. \cite{tian2023pyramid} develop a deep autoencoding Gaussian mixture model that integrates a pyramid reconstruction branch to capture multi-level features to robustly detect industrial faults.
Recently, much research has also been carried out in robotics anomaly detection. For example, Sindhwani et al. \cite{sindhwani2020unsupervised} develop an unsupervised anomaly detection framework for delivery drones, which trains a machine learning model to predict the flight dynamics and use the reconstruction loss to identify the anomalies during flying missions. Jiang et al. \cite{10599292} apply a spatio-temporal autoencoder to detect drone anomalies from flight data. 
In the field of manipulation robots, Yoo et al. \cite{yoo2021multimodal} propose an autoencoder-based network to detect whether the objects are slipped from the gripper of the robot using multi-modal data. Park et al. \cite{park2018multimodal} developed a Variational Autoencoder (VAE) based on LSTM for anomalous execution detection of robot-assisted feeding tasks.
% Sadhu et al. \cite{sadhu2023board} use a CRNN autoencoder with a classifier to detect and classify different anomalies of drones. 

However, attention is still underexplored on autonomous mobile robot anomaly detection. Most of the existing research still depends on prior knowledge of anomaly data for training and is restricted to detecting specific types of anomalies. For instance, Li et al. \cite{li2024novel} develop a supervised audio-visual network to detect vehicle collisions. Zhang et al. \cite{zhang2023fault} design a spatial-temporal graph attention network for hardware anomaly detection, which also follows a supervised manner. Therefore, there is a need for a robust unsupervised method capable of detecting various underlying anomalies for mobile robots.

% Kasap et al. \cite{kasap2023unsupervised} present an unsupervised framework using dissimilarity for anomaly detection in autonomous mobile robots. However, its effectiveness is limited to anomalies from metal bumpers, highlighting the need for a robust, unsupervised method capable of detecting various faults without relying on specific anomaly data.

\begin{figure*}[t]
    \centering
    \includegraphics[width=1.0\linewidth]{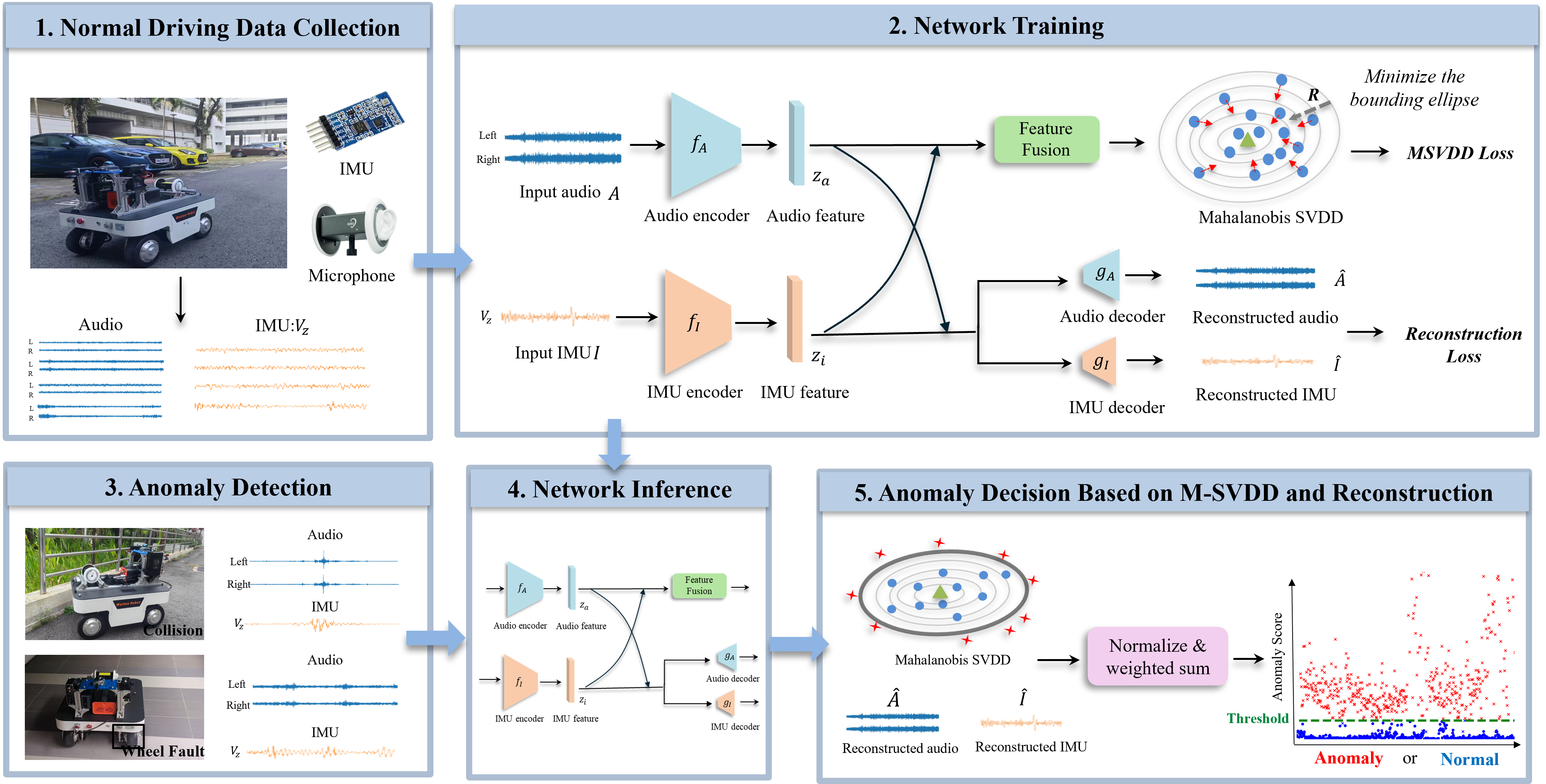}
    \caption{The overall workflow of the proposed unsupervised anomaly detection framework. In the training stage, audio and IMU data collected under normal conditions are input into the corresponding encoders for feature extraction. A cross-attention module is then used to fuse the features, which are fed into the M-SVDD branch to learn an ellipsoidal boundary enclosing normal data. Meanwhile, a reconstruction branch is applied to recover the original inputs, preventing feature collapse and improving detection accuracy. During inference, anomalies are detected based on reconstruction error and the Mahalanobis distance between the embedded feature and the center of the learned boundary.}
    \label{network}
\end{figure*}

\subsection{Support Vector Data Description}
% \begin{figure}[t]
%     \centering
%     \includegraphics[width=1\linewidth]{image/comparison.jpg}
%     \caption{Comparison between SVDD and GSVDD}
%     \label{svdd_gsvdd}
% \end{figure}
Support Vector Data Description (SVDD) \cite{tax2004support} is a classical method for anomaly detection, which aims to enclose normal data points within a hypersphere with a minimum radius. The objective function for SVDD is defined as:
\begin{equation}
    \begin{aligned}
        &\min_{R, c, \xi_i} \quad R^2 + C \sum_{i=1}^N \xi_i \\
        &\text{s.t.} \quad \| \phi(x_i) - c \|^2 \leq R^2 + \xi_i, \quad \xi_i \geq 0, \quad \forall i,
    \end{aligned}
\end{equation}
where $R$ and $c$ represent the radius and center of the hypersphere. $\phi(x_i)$ and $\xi_i$ denote the feature mapping of the input by a kernel function and slack variables for soft margin constraints, and $C$ is a regularization parameter. Due to its strong ability to model the boundary of normal samples without requiring prior knowledge of anomaly data, SVDD-based methods have been widely adopted in anomaly detection tasks. However, conventional SVDD-based methodologies encounter significant challenges when applied to high-dimensional datasets. Their dependence on kernel functions results in poor scalability due to the high computational cost of constructing and storing large kernel matrices. Moreover, SVDD operates on raw input features without learning high-level representations, reducing its effectiveness for complex inputs such as images, audio, or multimodal signals. 

Therefore, building on SVDD, Deep Support Vector Data Description (DSVDD) \cite{ruff2018deep} extends this approach by leveraging deep neural networks to learn a high-level latent representation of the data. The network maps input data $x_i$ to a latent space $\Phi(x_i;\mathcal{W})$, and the typical one-class DSVDD objective function is given by:
\begin{equation}
    \min_{\mathcal{W}} \frac{1}{n} \sum_{i=1}^n \left\| \Phi(x_i; \mathcal{W}) - \mathbf{c} \right\|_1 + \frac{\lambda}{2} \sum_{l=1}^L \left\| \mathcal{W}^l \right\|_F^2,
    \label{traditional svdd}
\end{equation}
where $\mathcal{W}$ denotes the network parameters, $\left\|. \right\|_F^2$ is the Frobenius norm and $\lambda$ is the hyperparameter controlling the weights of the regularization term.
Many DSVDD-based anomaly detection works have been proposed and applied to different fields. 
For example, Yi et al. \cite{yi2020patch} develop a patch-level SVDD to achieve accurate pixel-label anomaly detection for industrial images. Peng et al. \cite{peng2025reconstruction} developed a deep reconstruction-based unsupervised SVDD framework with an adaptive thresholding strategy for wind turbine anomaly detection. 
% Zhou et al. \cite{zhou2021vae} integrate VAE and SVDD and jointly optimize reconstruction loss and SVDD hypersphere to achieve better detection accuracy.

While DSVDD has proven to be a powerful tool for anomaly detection, it still has certain limitations that may reduce its effectiveness. First, DSVDD method faces the issue of hypersphere collapse. Without proper regularization, the network tends to achieve a trivial solution where all the network weights are zero to map all the features into the same point. Second, the use of Euclidean distance in SVDD implicitly assumes that all feature dimensions are independent and equally important, neglecting potential correlations and variations in feature scales. 

In this work, we introduce a M-SVDD to address the above limitations. By incorporating the covariance structure of the latent features, the model replaces the isotropic Euclidean metric with Mahalanobis distance, enabling adaptive scaling and correlation-aware boundary learning. Additionally, a reconstruction branch is introduced to preserve feature diversity and prevent representation collapse, thereby enhancing both the stability and accuracy of anomaly detection.

\section{Methodology}
The architecture of the proposed network is illustrated in Fig. \ref{network}. During the training phase, only normal driving audio and IMU data are fed into encoders for feature extraction. The extracted features are then processed in two parallel branches: a reconstruction branch to reconstruct the original signals and a M-SVDD branch, where the fused features are mapped into a latent space to enclose the normal data distribution. In the testing phase, anomalies are identified by evaluating their Mahalanobis distance from the hyperspace's center and the reconstruction loss. 
The overall workflow is presented in Algorithm \ref{alg:gsvdd}. This section gives a comprehensive overview of each module.

\subsection{Audio-IMU Feature Extraction}
The network accepts binaural audio $A\in \mathbb{R}^{L_A\times 2}$ and the velocity in $Z$ axis of IMU data $I\in \mathbb{R}^{L_I}$ as input to perceive the environmental sound and the vertical motion dynamics of the vehicle. $L_A$ and $L_I$ represent the length of the input audio and IMU signal. The audio and IMU data are input into specifically designed encoder $f(A;\theta_{AE})$ and $f(I;\theta_{IE})$ to obtain the corresponding feature $z_A\in \mathbb{R}^d$ and $z_I\in \mathbb{R}^d$, where $d$ represents the dimension of the feature.
In the paper, we adopt a CRNN network, which consists of six convolutional layers and an LSTM layer as an audio encoder to extract both long-range temporal dependencies and local features in audio sequences to identify anomalies like abrupt sounds or internal mechanical failure noises. While the IMU encoder incorporates two LSTM layers to extract the temporal dynamics of IMU data. 

The feature extracted from both modalities is then sent to a fusion module for information fusion. Specifically, a cross attention mechanism \cite{tao2021someone} as shown in Fig. \ref{cross-attention} is adopted for $z_A$ and $z_I$ to capture interactions between audio and IMU data:
\begin{equation}
z_{IA} = softmax(\frac{Q_{A}K_{I}^T}{\sqrt{d}})V_{I}, \\
z_{AI}= softmax(\frac{Q_{I}K_{A}^T}{\sqrt{d}})V_{A},
    \label{cross-attention}
\end{equation}
where $Q, K, V\in \mathbb{R}^d$ represents the query, key, and value vector of the feature, respectively. The cross-attention layer is then followed by a multilayer perceptron (MLP) layer and residual connection operation to obtain the final fusion feature $z_{IA}$ and $z_{AI}$ for audio and IMU data. By aligning key temporal features from audio and IMU signals, the model enhances its sensitivity to anomalies that simultaneously impact both modalities. Finally, $z_{AI}$ and $z_{IA}$ are summed and fed into an MLP layer to project the features into an $s$-dimensional space:
\begin{equation}
z = \mathcal{W}_M(z_{AI}+z_{IA})+b,
\label{equation:mapping}
\end{equation}
where $z \in \mathbb{R}^s$ is the fused audio-IMU feature, $\mathcal{W}_M\in \mathbb{R}^{s\times d}$ and $b \in \mathbb{R}^s$ denote the transformation matrix and bias term of the MLP layer.
\begin{figure}[t]
    \centering
    \includegraphics[width=0.95\linewidth]{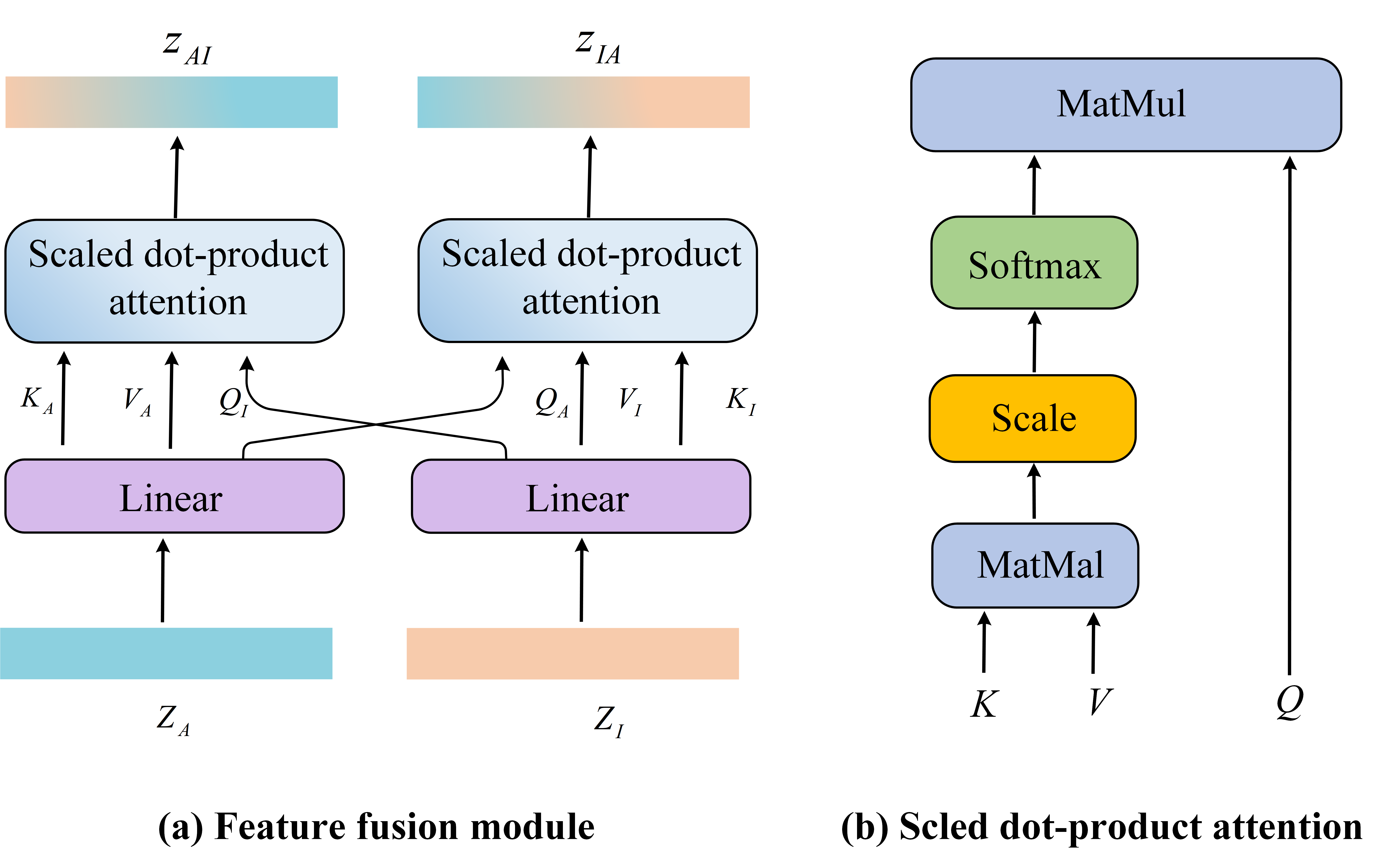}
    \caption{The cross-attention module for audio-IMU feature fusion.}
    \label{feature fusion}
    % \vspace{-3mm}
\end{figure}

\begin{algorithm}[t]
\caption{Workflow of the proposed method}
\label{alg:gsvdd}
\begin{algorithmic}[1]
\STATE \textbf{Input:} Binaural audio $A$, IMU signal $I$
\STATE \textbf{Initialize:} All model parameters $\theta^*$ and radius $R$
\STATE \textbf{Training Phase:}
\FOR{each training batch in dataset}
    \STATE Embedding: $z_{A}, z_{I} \leftarrow f(A;\theta_{AE}), f(I;\theta_{IE})$
    \STATE Reconstruction: $\hat{A}, \hat{I} \leftarrow g(z_A;\theta_{AD}), g(z_I;\theta_{ID})$
    \STATE Feature Fusion: $z \leftarrow$ using Eq.~\ref{cross-attention} and Eq.~\ref{equation:mapping}
    \STATE Estimate $\mu_z$, $\Sigma_z \leftarrow$ using Eq.~6 and Eq.~\ref{equation:u_calculation}
    \STATE Calculate Mahalanobis distance $D$ by Eq.~\ref{equation:mdistance}
    \STATE Calculate loss by Eq.~\ref{equation:total_loss}
    \STATE Update model parameters and $R$
\ENDFOR
\STATE \textbf{Inference Phase:}
\STATE Obtain anomaly score $\delta$ for each training sample by \\ Eq.~\ref{equaion:anomaly_score}
\STATE Threshold $\delta^* \leftarrow$ 95th percentile of all training 
\\anomaly scores
\STATE Compare $\delta_{test}$ with $\delta^*$, if $\delta_{test} >\delta^*\rightarrow$ anomaly.
\end{algorithmic}
\end{algorithm}
\subsection{Mahalanobis SVDD}
The objective of the M-SVDD module is to project the fused features $z$ of normal data into a compact hyperspace with a minimum radius, where anomalous samples are expected to be positioned outside the bounding ellipse and at a significant distance from the center. 
Typically, this is achieved by Eq. \ref{traditional svdd}, where the features are reflected into a bounding sphere, and Euclidean distance is utilized to measure the distance between the feature and the center. However, it does not take the correlation between different feature dimensions into account and assumes that all features contribute equally in determining the anomaly boundary, and it may be hard to describe the normal data if its distribution is complex and has many outliers. 
To address these limitations, we replace the Euclidean distance with the Mahalanobis distance, which accounts for both the variance and the inter-dependencies among feature dimensions. The anomaly score is computed as: 
\begin{equation}
    D_i = \sqrt{(z_i - \mu_z)^T \Sigma_z^{-1} (z_i - \mu_z)},
    \label{equation:mdistance}
\end{equation}
where $\mu_z\in \mathbb{R}^s$ and $\Sigma_z\in \mathbb{R}^{s\times s}$ are the mean and the covariance matrix of the features for the training samples. Unlike Euclidean distance,  Mahalanobis distance incorporates the inverse of the feature covariance matrix $\Sigma_z^{-1}$, which adjusts the contribution of each feature dimension based on its variance and captures the correlations between them. As a result, directions with high variance (less informative) are down-weighted, while those with low variance (more informative) are emphasized. This allows the model to form an anisotropic, ellipsoidal boundary that better aligns with the actual distribution of normal data.
% Instead of using Euclidean distance, the application of Mahalanobis distance enables the network to adaptively adjust the importance of each feature and consider the correlation between different features by analyzing the covariance $\Sigma_z$. Moreover, 
% Compared with the sphere decision boundary of Euclidean distance-based SVDD, the GSVDD is capable of generating an elliptical decision boundary rather than a rigid spherical one to better represent the distribution of normal data. 

To accurately compute the Mahalanobis distance, reliable estimates of the mean and covariance of the latent features are essential. However, although training data are collected under normal operating conditions, outliers caused by noise or transient disturbances remain inevitable. These outliers can severely skew the estimation of $\mu_z$ and $\Sigma_z$, degrading detection performance.
To address this challenge, the Minimum Covariance Determinant (MCD) estimator \cite{0d23955b-2c3d-3692-b621-b91e68c0944a} is employed for determining $\mu_z$ and $\Sigma_z$. 
% The MCD estimator is specifically designed to mitigate the influence of outliers and improve numerical stability by avoiding issues such as the amplification of small eigenvalues.
Specifically, given a batch of $N$ training samples, we select a subset $H$ of size $h < N$ whose sample covariance matrix has the minimum determinant within the batch.
These selected $h$ samples are considered to capture the most compact and representative region of the normal data distribution and are then utilized to estimate $\mu_z$ and $\Sigma_z$:
\begin{align}
    \mu_z &= \frac{1}{h} \sum_{i \in H} x_i, \\
    \Sigma_z &= \frac{1}{h} \sum_{i \in H} (z_i - \mu_z)(z_i - \mu_z)^T+\epsilon I,
    \label{equation:u_calculation}
\end{align}
where $\epsilon I$ is a regularization matrix to avoid the singularity issue. $I$ is the identity matrix, and $\epsilon$ is a hyperparameter, which is set to 0.001 in the paper. 
Subsequently, soft boundary loss \cite{ruff2018deep} is employed to learn a minimal enclosing ellipsoidal boundary that captures the distribution of normal samples while reduce the influence of outliers. The objective function for the Mahalanobis SVDD is defined as follows:
\begin{equation}
    {\mathcal{L}_{MSVDD}} = R^2+ \frac{1}{N}\sum_{i=1}^{N}Max(0,D_i^2-R^2),
    \label{equation:GSVDD_loss}
\end{equation}
where $R$ is the radius of the learned ellipsoidal space and is a learnable parameter. 
% \begin{equation}
%     D_i = \sqrt{(z_i - \mu_z)^T \Sigma_z^{-1} (z_i - \mu_z)}.
%     \label{equation:mdistance}
% \end{equation}
% Instead of using Euclidean distance to measure the distance, the application of Mahalanobis distance enables the network to adaptively adjust the importance of each feature and consider the correlation between different features by analyzing the covariance $\Sigma_z$. Moreover, compared with the sphere decision boundary of Euclidean distance-based SVDD, the GSVDD is capable of generating an elliptical decision boundary rather than a rigid spherical one, which is essential for representing the true distribution of the normal data. 

\subsection{Reconstruction and Regularization} 
A reconstruction branch is introduced into the network, where $z_A$ and $z_I$ are passed through dedicated decoders to reconstruct the original input signals. This branch offers two key advantages. First, DSVDD-based methods commonly suffer from the issue of hypersphere collapse, where network weights tend to converge to zero in an attempt to minimize the hypersphere's radius, resulting in the projection of all samples onto a singular feature. By incorporating the reconstruction task, the network is compelled to learn diverse feature representations, thereby mitigating the collapse issue \cite{peng2025reconstruction}. Additionally, since the network is only trained on normal operation data, it tends to fail to accurately reconstruct anomalous inputs, leading to a significantly higher reconstruction loss for anomaly samples, while normal data yields a relatively low loss. As a result, reconstruction loss also serves as an essential metric for effective anomaly detection.

In this work, the decoders are designed to follow a symmetric structure with respect to the encoders. Specifically, the audio decoder $g(z_A;\theta_{AD})$ consists of six deconvolutional layers to progressively restore fine-grained temporal information from the latent representation, while the IMU decoder $g(z_I;\theta_{ID})$ is based on LSTM layers. The networks aim to minimize the reconstruction loss for both audio and IMU signals:
\begin{equation}
    \mathcal{L}_{{Rec}} = \mathcal{L}_{Huber}(A,\hat{A})+\mathcal{L}_{Huber}(I,\hat{I}),
    \label{equation:reconstruction_loss}
\end{equation}
where $\hat{A}_i\in \mathbb{R}^{L_A\times 2}$ and $\hat{I}_i\in \mathbb{R}^{L_I}$ are the reconstructed audio and IMU signal output by the decoders, and $\mathcal{L}_{Huber}$ is given by:
\begin{equation}
\mathcal{L}_{Huber}(y, \hat{y}) =
\begin{cases}
    \frac{1}{2}(y - \hat{y})^2 & \text{for } |y - \hat{y}| \leq 1, \\
    |y - \hat{y}| - \frac{1}{2}\ & \text{otherwise}.
\end{cases}
\label{huber_loss}
\end{equation}

% To further ensure the network avoids converging to suboptimal local minima and to prevent high-dimensional collapse, a regularization term is introduced to increase the information entropy, encouraging a broader distribution of the extracted feature $z$. The information entropy for a multivariate Gaussian distribution is given by $\frac{1}{2}\ln((2\pi e)^d|\Sigma|)$. Thus, the corresponding regularization loss is formulated as:
Furthermore, to enhance the numerical stability of matrix operations and improve feature separation, a regularization loss is introduced to the network to penalize the small diagonal entries of the covariance matrix and large conditional numbers:
\begin{equation}
%     \mathcal{L}_{Reg} = -ln|\Sigma_z|.
    \mathcal{L}_{Reg} = \sum_{i=1}^{s}\left( \frac{1}{\sigma_i + \epsilon}\right) + \frac{\lambda_{\max}}{\lambda_{\min} + \epsilon},
    \label{equation:reg_loss}
\end{equation}
where $\sigma$, $\lambda_{\max}$ and $\lambda_{\min}$ represent the diagonal value, the largest and the smallest eigenvalue of $\Sigma_z$.

The overall loss function for the framework is given by:
\begin{equation}
    \mathcal{L}_{total} = \alpha_1\mathcal{L}_{MSVDD}+\alpha_2\mathcal{L}_{{Rec}}+\alpha_3\mathcal{L}_{Reg},
    \label{equation:total_loss}
\end{equation}
where $\alpha_1$, $\alpha_2$ and $\alpha_3$ are hyperparameters. Note that the combination weights parameters are manually set in the current version, based on empirical tuning. In future work, these can be learned via reinforcement learning or differentiable optimization modules for full end-to-end training.

\subsection{Network Inference} 
The inference stage for anomaly detection utilizes both the reconstruction loss and the SVDD loss as the criterion. The final anomaly score $\delta$ for the input sample is given by:
\begin{equation}
\delta_i = D_i + w\frac{\mu_{D}^{\mathcal{T}}}{\mu^{\mathcal{T}}_{\mathcal{L}_{Rec}}}\mathcal{L}_{Rec}^i,
\label{equaion:anomaly_score}
\end{equation}
where $\mu^{\mathcal{T}}_{D}$ and $\mu^{\mathcal{T}}_{\mathcal{L}_{Rec}}$ represents the mean distance and reconstruction loss on the training dataset, ensuring that both metrics are normalized to the same scale, while $w$ is the hyperparameter to control the weights between two metrics. The anomaly threshold $\delta^*$ is defined as the 95th percentile of the anomaly scores across all training samples. A test sample is classified as anomalous if its anomaly score exceeds the threshold.

\begin{figure}[t]
    \centering
    \includegraphics[width=0.9\linewidth]{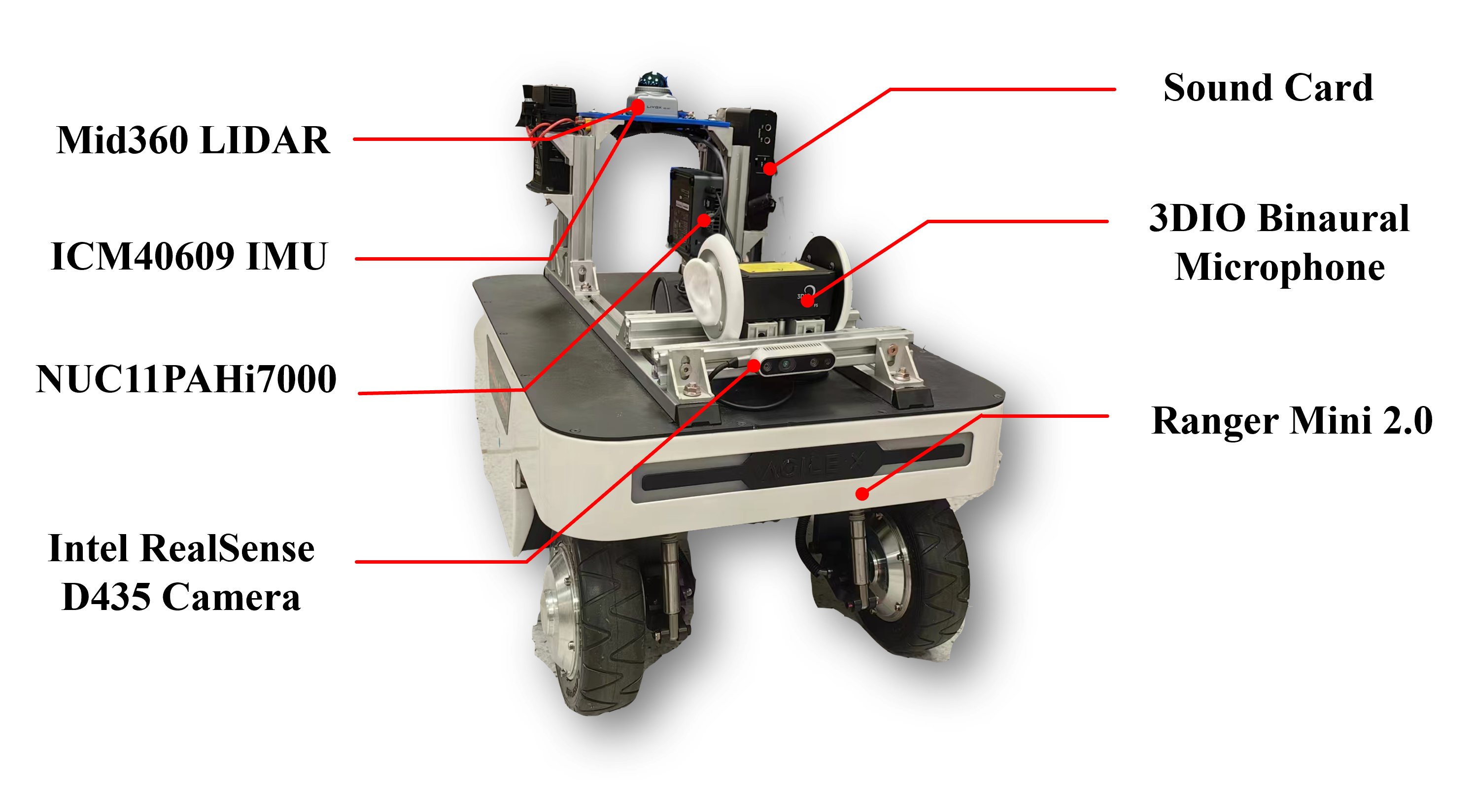}
    \caption{The experimental platform utilized for data collection.}
    \label{dataset}
    % \vspace{-3mm}
\end{figure}

% \begin{table}[t]
% \centering
% \renewcommand{\arraystretch}{1.3}
% \setlength{\tabcolsep}{7pt} 
% \caption{The number of collected samples for each type used in training and testing.}
% \begin{tabular}{ccc}
% \hline
% &\textbf{\# Training sample} & \textbf{\# Test sample} \\
% \hline
% Normal-Indoor & 2523 & 642 \\
% Normal-Outdoor & 2564& 574 \\
% Anomaly-Collision & 0 & 279 \\
% Anomaly-Machinery & 0 & 513 \\
% \hline
% \end{tabular}
% \label{data split}
% \end{table}

\section{Experiments}
\begin{figure}[t]
    \centering
    \includegraphics[width=0.95\linewidth]{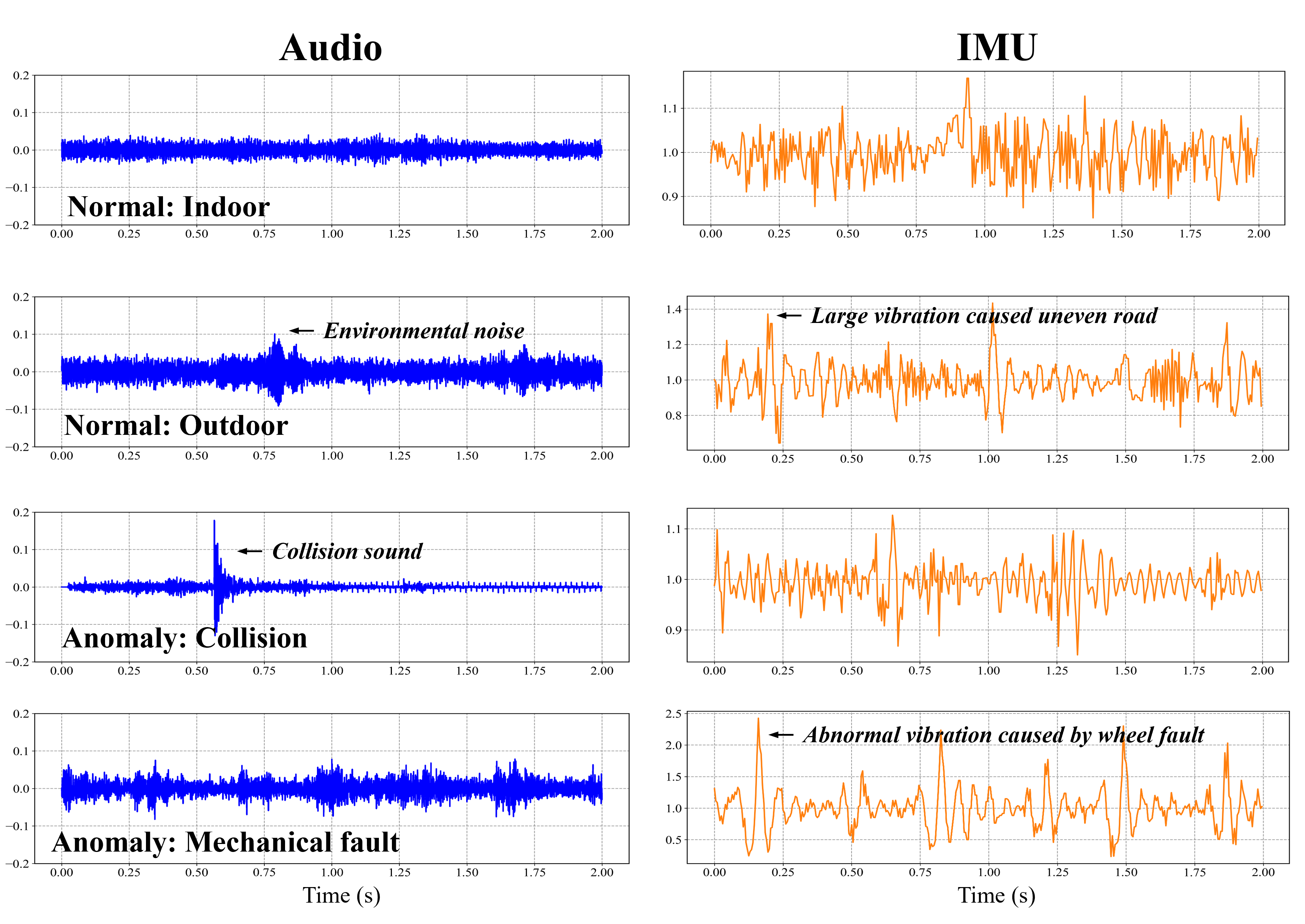}
    \caption{Examples of selected samples from normal driving and anomalous signals.}
    \label{data_samples}
    \vspace{-3mm}
\end{figure}

To evaluate the effectiveness of the proposed method, we have created a multi-modal anomaly detection dataset specifically for autonomous mobile robots, which includes image, audio, LiDAR, and IMU data collected using the platform illustrated in Fig. \ref{dataset}.
The dataset comprises two hours of normal driving data captured in both indoor and outdoor environments, encompassing diverse road surfaces, visual contexts, and background soundscapes. Two types of abnormal scenarios are included in the anomaly data. The first involves instances where the autonomous robot unexpectedly collides with surrounding objects such as doors, walls, curbs, desks, chairs and pedestrians. The second type captures data under internal vehicle malfunctions, specifically simulating the performance of the robot under various wheel faults. 
Examples of the collected data are presented in Fig. \ref{data_samples}.
% with corresponding driving data recorded for these scenarios. Some of the collected data are presented in Fig. \ref{sample_vis}.

\subsection{Experimental Setting} 
\textbf{Data preparation}: To preprocess the data for the experiments, all audio and IMU sequences are segmented into 2-second intervals. The audio signals are downsampled by a factor of 10, resulting in a final sampling rate of 4410 Hz to reduce computational complexity. 
Only the Z-axis velocity from IMU data is used, as it captures more representative anomaly features. 
While multi-axis data may improve performance, access to all inertial channels is not always guaranteed in practice. This setting allows us to evaluate the robustness of the proposed method under limited sensor input.
The data is then standardized using Z-score normalization, followed by the application of a smoothing filter to enhance its quality and reduce noise. The IMU data is collected at 200Hz. Therefore, $L_A$ and $L_I$ are set to 8820 and 400 in the experiments. There are a total of 21 normal driving sequences in the dataset. We select 15 of them for training and the remaining for testing, while all the anomaly data is only used for testing.
% The detailed information of the dataset division is presented in Table \ref{data split}.

\textbf{Evaluation metrics}: We utilize precision, recall and F1 score (F1) as evaluation protocols:
\begin{equation}
    P = \frac{TP}{TP + FP}, \quad R = \frac{TP}{TP + FN}, \quad F1 = 2 \cdot \frac{P \cdot R}{P + R},
\end{equation}
where $TP$, $FP$, and $FN$ represent true positive, false positive, and false negative, respectively. Since some of the comparison methods do not explicitly define how to set the decision threshold, we follow prior works \cite{xu2024calibrated,audibert2020usad,li2021multivariate}, by selecting the optimal threshold with the best F1-score to ensure a fair comparison. 

\textbf{Implementation details}: 
The network is trained using the Adam optimizer with a batch size of 64 over 40 epochs. The initial learning rate is set to $1 \times 10^{-4}$ and is updated throughout training using a cosine decay schedule.
The hyperparameters $\alpha_1, \alpha_2$, and $\alpha_3$ are assigned the values of 1, 1, 0.001, respectively. The dimensions of the feature and ellipsoidal latent space $d$ and $s$ are set to 32, while $w$ is set to 0.01 to decide the anomaly threshold. All the experiments are performed on a workstation equipped with an NVIDIA GeForce RTX 4080 GPU.

\subsection{Comparison Experiments} 
We conduct a comprehensive comparison of the proposed method with various baseline methods, including traditional shallow machine learning approaches, which depend on handcrafted features and simple statistical models, and deep learning techniques that learn hierarchical representations from multimodal data through neural networks. These selected baselines are general-purpose and modality-agnostic methods that are not specifically designed for any particular sensor type or data domain~\cite{yi2020patch,carratu2023novel,xu2024calibrated,li2024novel}. Most of them are publicly available with open-source implementations, supporting reproducibility and fair evaluation. In particular, we focus on methods that do not require labeled anomaly data, aligning with the unsupervised nature of our task. Specifically, we include:

% \begin{table}[t]
% \centering
% \renewcommand{\arraystretch}{1.3}
% \setlength{\tabcolsep}{9pt}  % Adjust column spacing
% \caption{The AUC scores (\%) for each collision and machinery type, as well as the overall AUC for detecting both anomalies.}
% \begin{tabular}{l|ccc}
% \hline
% \textbf{Methods} & \textbf{Collision} & \textbf{Machinery} & \textbf{Overall} \\
% \hline
% IF \cite{liu2008isolation} & 79.7  & 96.5  & 90.5  \\
% OCSVM \cite{manevitz2001one} & 91.8  & \underline{96.7}  & 95.0  \\
% SVDD \cite{tax2004support} & 91.7  & \textbf{96.8}  & 95.0 \\
% % UDFD \cite{kasap2023unsupervised}  & 97.0  & 83.3  & 89.6  \\
% DeepForest \cite{xu2023deep} & 91.0  & 96.6  & 94.7  \\
% DAGMM \cite{zong2018deep}  & \underline{98.3}  & 93.2  & {95.2}  \\
% DAE &  93.8 &  94.4 &  95.3 \\
% VAE-SVDD  \cite{zhou2021vae} & 97.4  & 94.0  & 95.2  \\ 
% DSVDD  \cite{ruff2018deep} & 96.4  & 95.2 & \underline{95.6}  \\
% \hline
% \textbf{Ours} & \textbf{98.5} & 95.5 & \textbf{97.0} \\
% \hline
% \end{tabular}
% \label{comparison_results_auc}
% \end{table}

\begin{table*}[t]
\centering
\renewcommand{\arraystretch}{1.6} % Adjust column spacing
\setlength{\tabcolsep}{9pt} 
\caption{Comparative analysis of different methods. The best results are in \textbf{bold} while the second-best results are \underline{underlined}. In addition to standard metrics, we also report the overall AUC as well as class-specific AUCs for collision-related (AUC\textsubscript{coll}) and mechanical-failure-related (AUC\textsubscript{mech}) anomalies to provide a more detailed evaluation.}
\begin{tabular}{lccccccc}
\hline
\textbf{Methods} & \textbf{Type} &\textbf{Recall(\%)} & \textbf{Precision(\%)} & \textbf{F1(\%)} & \textbf{AUC(\%)}&\textbf{AUC\textsubscript{coll}(\%)} & \textbf{AUC\textsubscript{mech}(\%)} \\
\hline
IF \cite{liu2008isolation}&Shallow  & 80.9  & 81.2  & 81.1& 90.5  & 79.7  & 96.5   \\
OCSVM \cite{manevitz2001one}&Shallow  & 89.0  & {87.5}  & 88.2 & 95.0 & 91.8  & \underline{96.7}   \\
SVDD \cite{tax2004support}&Shallow  & 89.6  & 87.0  & 88.3  & 95.0& 91.7  & \textbf{96.8} \\
% UDFD \cite{kasap2023unsupervised}  & 97.0  & 83.3  & 89.6  \\
DeepForest \cite{xu2023deep}&Deep  & 89.3  & 86.9  & 88.0 & 94.7 & 91.0  & 96.6  \\
DAGMM \cite{zong2018deep} &Deep & 97.1  & 86.3  & \underline{91.4}  & {95.2} & \underline{98.3}  & 93.2 \\
DAE&Deep  & 91.9  & \underline{87.7} & 89.8 &  95.3 &  93.8 &  94.4  \\
% Autoencoder &  &   &   \\
VAE-SVDD  \cite{zhou2021vae}&Deep  & 94.3  & 84.8  & 89.3 & 95.2 & 97.4  & 94.0    \\ 
DSVDD  \cite{ruff2018deep} &Deep & 96.0  & 87.0 & 91.2 & \underline{95.6} & 96.4  & 95.2 \\
\hline
\textbf{Ours}&Deep & \textbf{97.5} & \textbf{87.7} & \textbf{92.3}& \textbf{97.0} & \textbf{98.5} & 95.5 \\
\hline
\end{tabular}
\label{comparison_results}
\end{table*}

% \begin{table}[t]
%     \centering
%     \caption{The contribution of each module to the overall performance metrics.}
%     \renewcommand{\arraystretch}{1.3} 
%     \setlength{\tabcolsep}{5pt}       
%     \begin{tabular}{c|ccc}

%         \hline
%         \textbf{Method} & \textbf{Recall(\%)} & \textbf{Precision(\%)} & \textbf{F1(\%)} \\ 
%         \hline
%         \ \textbf{w\slash o Audio} & 64.4 & \textbf{91.2} & 75.4 \\
%         \ \textbf{w\slash o IMU }& 96.3 & 87.6 & 91.8 \\
%         \ \textbf{w\slash o Reconstruction} & 97.2 & 81.7 & 88.8 \\
%         \ \textbf{w\slash o M-SVDD} & 91.9  & 87.7  & 89.8  \\ \hline
%         \ \textbf{Ours} & \textbf{97.5} & 87.7 & \textbf{92.3} \\
%     \hline
%     \end{tabular}
%     \label{ablation:module}
%     \vspace{-2mm}
% \end{table}

\begin{table*}[t]
    \centering
    \renewcommand{\arraystretch}{1.4} % Adjust row height
    \setlength{\tabcolsep}{12pt}    
    \caption{The contribution of each module to the overall performance metrics.}
    \begin{tabular}{cccc|ccc}
        \hline
        \textbf{Audio} & \textbf{IMU}& \textbf{Reconstruction}& \textbf{M-SVDD} & \textbf{Recall(\%) } & \textbf{Percision(\%) } & \textbf{F1(\%) } \\ \hline
        \xmark  & \cmark & \cmark& \cmark & 64.4 & \textbf{91.2} & 75.4 \\
        \cmark &\xmark & \cmark& \cmark  & 96.3 & 87.6 & \underline{91.8}\\
        \cmark & \cmark & \xmark& \cmark&  \underline{97.2} & 81.7 & 88.8 \\
        \cmark &\cmark & \cmark& \xmark   & 91.9  & 87.7  & 89.8 \\
        \cmark & \cmark & \cmark& \cmark& \textbf{97.5} & \underline{87.7} & \textbf{92.3}\\
        \hline
    \end{tabular}
    \label{ablation:module}
\end{table*}
% \begin{table}[t]
% \centering
% \renewcommand{\arraystretch}{1.3}
% \setlength{\tabcolsep}{9pt}  % Adjust column spacing
% \caption{Comparative analysis of different methods. The best results are in \textbf{bold} while the second-best results are \underline{underlined}.}
% \begin{tabular}{l|ccc}

% \hline
% \textbf{Methods} & \textbf{Recall(\%)} & \textbf{Precision(\%)} & \textbf{F1(\%)} \\
% \hline
% IF \cite{liu2008isolation} & 80.9  & 81.2  & 81.1  \\
% OCSVM \cite{manevitz2001one} & 89.0  & {87.5}  & 88.2  \\
% SVDD \cite{tax2004support} & 89.6  & 87.0  & 88.3 \\
% % UDFD \cite{kasap2023unsupervised}  & 97.0  & 83.3  & 89.6  \\
% DeepForest \cite{xu2023deep} & 89.3  & 86.9  & 88.0  \\
% DAE & 91.9  & \underline{87.7}  & 89.8  \\
% DAGMM \cite{zong2018deep}  & 97.1  & 86.3  & \underline{91.4}  \\
% % Autoencoder &  &   &   \\
% VAE-SVDD  \cite{zhou2021vae} & 94.3  & 84.8  & 89.3  \\ 
% DSVDD  \cite{ruff2018deep} & 96.0  & 87.0 & 91.2  \\
% \hline
% \textbf{Ours} & \textbf{97.5} & \textbf{87.7} & \textbf{92.3} \\
% \hline
% \end{tabular}
% \label{comparison_results}
% \end{table}

\begin{figure}[t]
    \centering
    \includegraphics[width=0.98\linewidth]{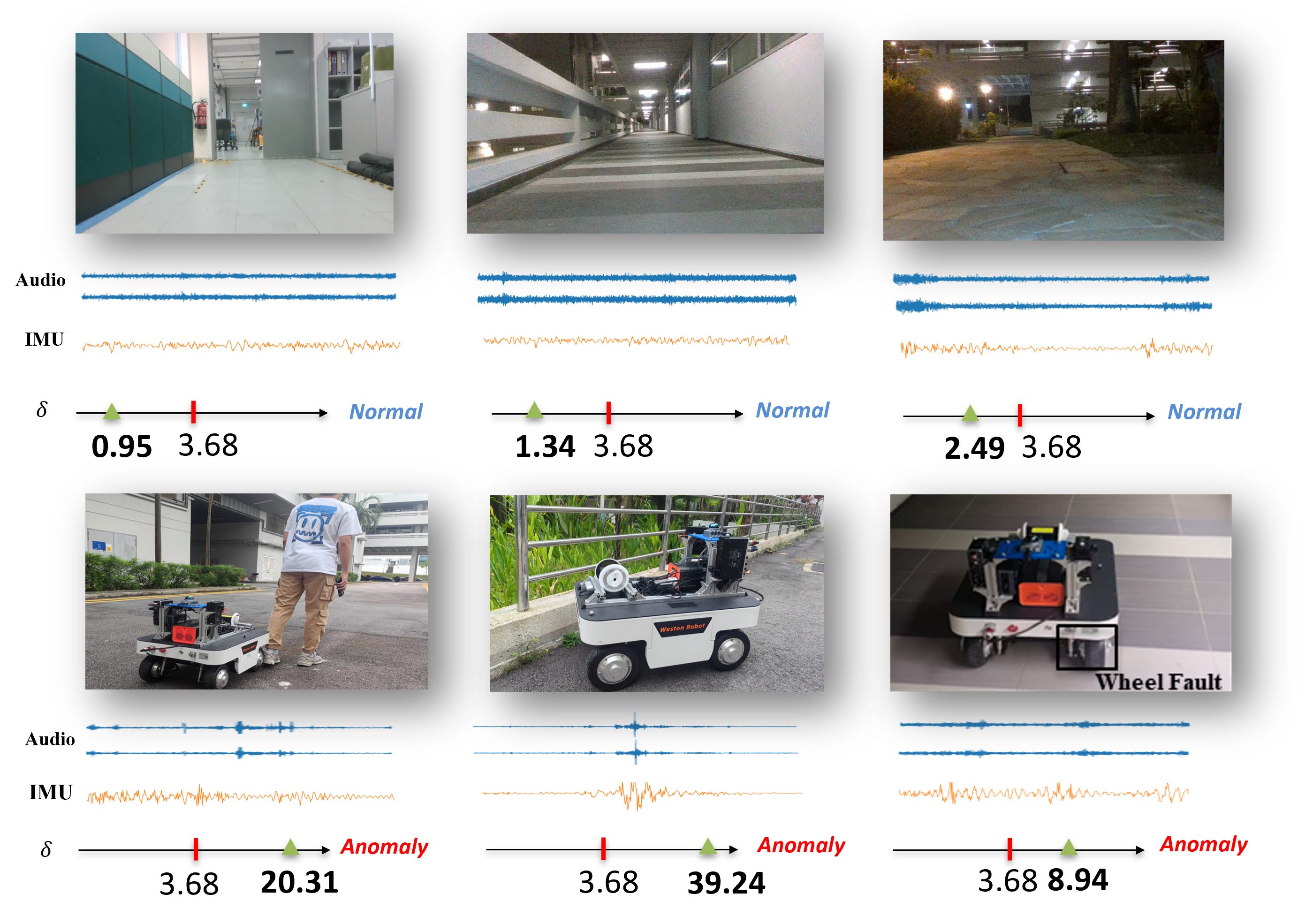}
    \caption{The inference results of test samples in the dataset.}
    \label{sample_vis}
    \vspace{-2mm}
\end{figure}

\begin{itemize}
    \item Isolation Forest (IF)\footnote{\url{https://scikit-learn.org}}\cite{liu2008isolation}, One-Class Support Vector Machine (OCSVM)\footnotemark[1]\cite{manevitz2001one} and SVDD\footnote{\url{https://github.com/KaitaiZhang/SVDD}}\cite{tax2004support}, which detect anomalies by random feature selection, separating data using high-dimensional hyperplanes, and enclosing normal samples within a minimum-radius hypersphere, respectively.
    % \item UDFD \cite{kasap2023unsupervised}: This method calculates the dissimilarity between different time sequence windows to identify the anomaly.
    \item DeepForest\footnote{\url{https://github.com/xuhongzuo/DeepOD}}\cite{xu2023deep}: This method extends IF, which utilizes a neural network to reflect the input data into different feature clusters.
    \item DAGMM\footnote{\url{https://github.com/danieltan07/dagmm}}\cite{zong2018deep}: It uses an autoencoder to extract a low-dimensional feature from the input, which is then input into a Gaussian Mixture Model to represent the distribution of the normal data.
    \item DAE: The deep autoencoder (DAE) reconstructs the input signals and uses the reconstruction loss for anomaly detection.
    \item VAE-SVDD \cite{zhou2021vae}: This model utilizes a Variational autoencoder (VAE) for signal reconstruction, and the encoded feature of VAE is employed for SVDD.
    \item DSVDD\footnotemark[3] \cite{ruff2018deep}: The same feature extraction and fusion module in the proposed method is used for DSVDD to map the data into a hypersphere, while the reconstruction branch is also incorporated \cite{peng2025reconstruction}.
\end{itemize}
Since traditional machine learning methods are not inherently designed to handle multi-modal input, we train these models on both early fusion and late fusion strategies, and the best-performing results are selected. For deep-learning methods, we utilize the same encoder, decoder and fusion module to ensure a fair comparison.
%to ensure a fair and consistent comparison across methods. 

% \begin{figure*}[t]
%     \centering
%     \subfloat[]{%
%         \raisebox{0mm}{\includegraphics[width=0.30\linewidth]{image/roc_curve_high_res.png}}%
%         \label{fig:mach}
%     }
%     \hfill
%     \subfloat[]{%
%         \raisebox{0mm}{\includegraphics[width=0.30\linewidth]{image/roc_curve_high_res.png}}%
%         \label{fig:coli}
%     }
%     \hfill
%     \subfloat[]{%
%         \raisebox{0mm}{\includegraphics[width=0.30\linewidth]{image/roc_curve_high_res.png}}%
%         \label{fig:overall}
%     }
%     \caption{Figures (a) and (b) illustrate the variance distribution in the PCA space for DSVDD and GSVDD, respectively, while Figure (c) depicts the covariance matrix of the features for both normal and anomalous data.}
%     \label{pr-curve}
% \end{figure*}

The comparison results are presented in Table \ref{comparison_results}. Our method achieves the highest F1-score (92.3\%), recall (97.5\%) and precision (87.7\%), outperforming both shallow machine learning and deep learning-based approaches.
Specifically, traditional methods struggle to detect the anomalies and achieve a relatively low recall and F1 score. For example, the recall for SVDD is only 89.6\%, which is 7.9\% lower than that of the proposed method. Among the deep learning models, DAGMM and DSVDD deliver competitive recall values of 97.1\% and 96.0\%, but their lower precision results in less balanced performance, as indicated by lower F1-scores. 

\begin{figure}[t]
    \centering
    \includegraphics[width=0.9\linewidth]{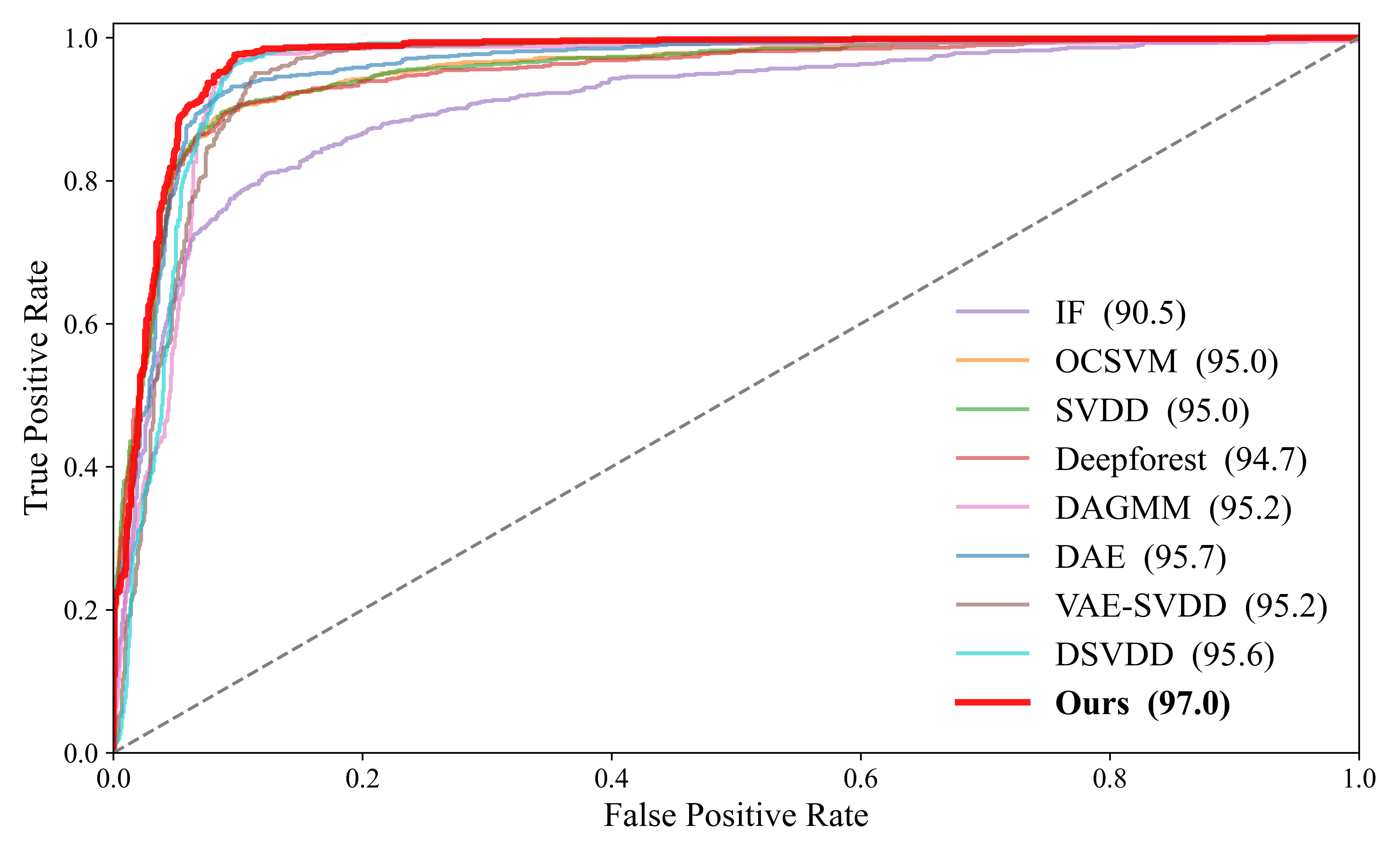}
     \caption{The ROC curves for different methods. The number in the bracket denotes the AUC score (\%).}
    \label{pr-curve}
    % \vspace{-4mm}
\end{figure}

The Receiver Operating Characteristic (ROC) curves for different comparison methods are plotted in Fig. \ref{pr-curve}. It shows that our method outperforms other approaches, obtaining the highest Area Under Curve (AUC) of 97.0\% across a wide range of threshold values. Deep learning models like VAE-SVDD and DSVDD also demonstrate competitive performance, achieving around 95.2\%-95.6\% AUC. But they exhibit relatively lower true positive rate (TPR) when false positive rate (FPR) is constrained to low values. In contrast, traditional methods perform significantly worse, with IF achieving the lowest AUC score of 90.5\%. The results further highlight that our method is able to effectively balance accurate anomaly detection with minimal false alarms, which is essential in real-world anomaly detection scenarios where reducing false positives is crucial. 

Then, we separately test the performance of the comparison methods on each machine type. The results in Table \ref{comparison_results} show that our method is able to detect both types of faults well under different FPRs. While shallow machine learning methods achieve better performance on machinery fault detection, they fail to detect collision anomalies. In contrast, our proposed method achieves the highest AUC score on collision anomalies and also performs well on machinery faults, obtaining the highest AUC score among deep learning methods.

Fig. \ref{sample_vis} illustrates some representative inference results on test samples from the dataset. As shown, the proposed network outputs high anomaly scores to different anomalous events, while consistently producing low scores during normal driving conditions.

\begin{figure*}[t]
    \centering
    \subfloat[]{%
        \raisebox{0mm}{\includegraphics[width=0.22\linewidth]{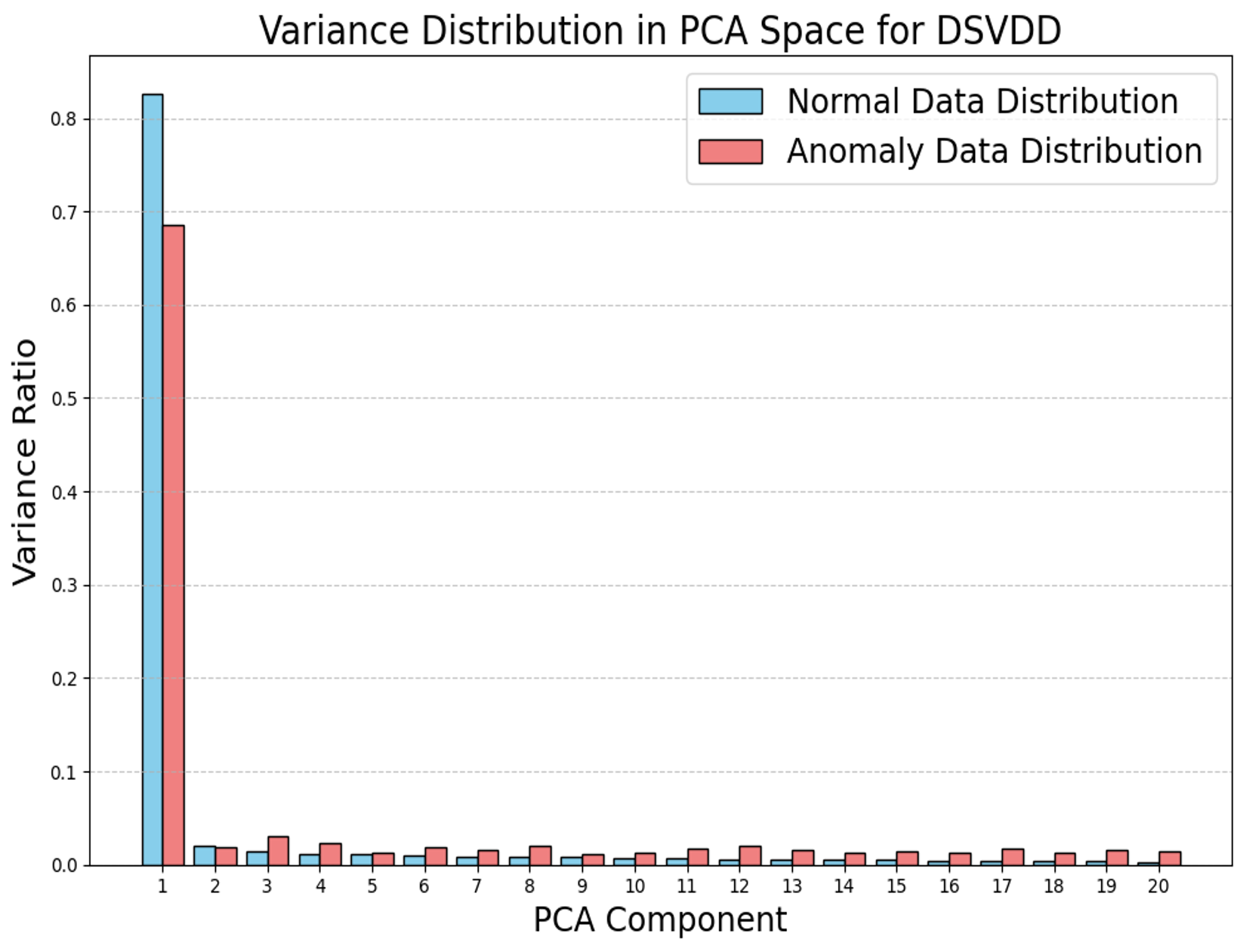}}%
        \label{fig:sub1}
    }
    \hfill
    \subfloat[]{%
        \raisebox{0mm}{\includegraphics[width=0.22\linewidth]{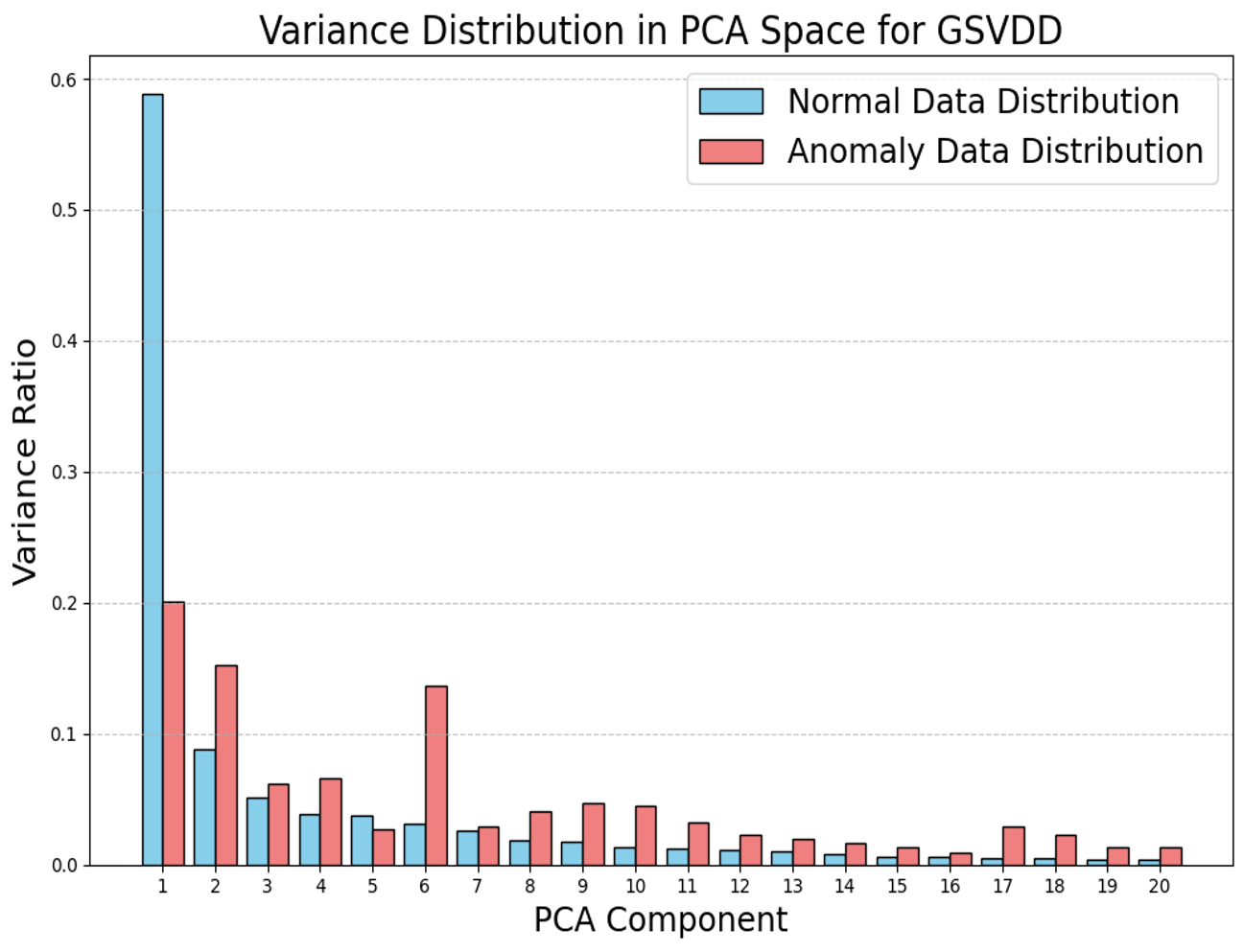}}%
        \label{fig:sub2}
    }
    \hfill
    \subfloat[]{%
        \raisebox{0.1cm}{\includegraphics[width=0.49\linewidth]{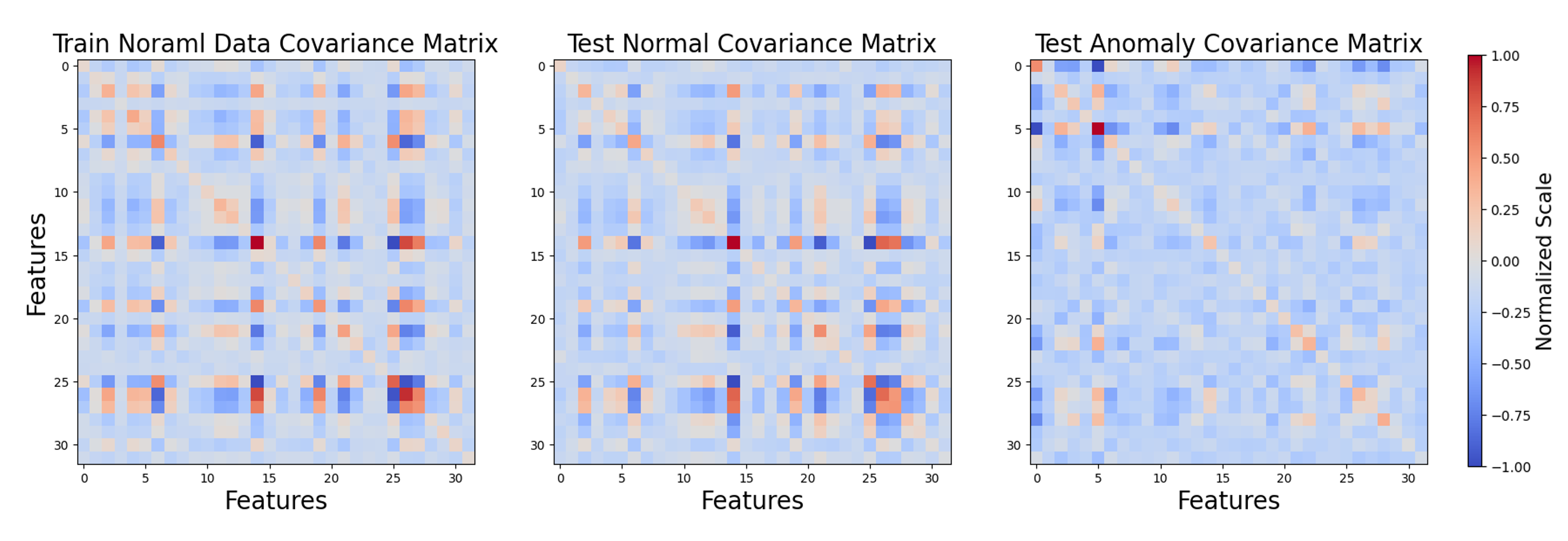}}%
        \label{fig:sub3}
    }
    \caption{Figures (a) and (b) illustrate the variance distribution in the PCA space for DSVDD and M-SVDD, respectively, while Figure (c) depicts the covariance matrix of the features for both normal and anomalous data of M-SVDD.}
    \label{fig:main}
    % \vspace{-2mm}
\end{figure*}

\subsection{Ablation Study}
Ablation experiments are conducted to evaluate the effectiveness of each modality and module. The results in Table \ref{ablation:module} indicate that removing the audio modality causes a significant drop in recall from 97.5\% to 64.4\%, though precision remains relatively high at 91.2\%. This demonstrates the critical role of audio data in detecting anomalies and contributing to the overall performance. While the integration of IMU data also plays a supportive role in anomaly detection, without IMU input, the recall and F1-score decrease by 1.2\% and 0.7\%, respectively.

Then, We evaluate the effectiveness of the proposed M-SVDD and reconstruction module. Excluding the M-SVDD module leads to a noticeable drop in recall to 91.9\%. This demonstrates that the application of the M-SVDD module enables effective detection for various complex anomalies. 
In contrast, the reconstruction module proves essential for maintaining high precision, as its removal causes precision to fall sharply to 81.7\% and F1 score drop significantly to 88.8\%. Without the reconstruction branch, minimizing the hyperspace radius may result in degenerate representations, as observed in previous SVDD-based study \cite{ruff2018deep}. The complete model, incorporating both modules, achieves the best performance with an F1-score of 92.3\%, showing that both M-SVDD and reconstruction branches are crucial for maintaining a balance between recall and precision.

Furthermore, we test the effectiveness of the cross-attention feature fusion module. The results presented in Table~\ref{ablation:crossattention} show that, compared with direct concatenation, element-wise summation, and product operations, the cross-attention strategy better integrates useful information from both modalities. Specifically, it achieves the highest F1 score, which is 2.0\% higher than that of concatenation and element-wise product, and 0.6\% higher than the summation method. This improvement is likely due to the ability of cross-attention to focus on shared and relevant features across modalities, which helps enhance the representation and improve detection performance.
\begin{table}[t]
    \centering
    \caption{The effectiveness of the cross-attention feature fusion module.}
    \renewcommand{\arraystretch}{1.5} 
    \setlength{\tabcolsep}{6pt}       
    \begin{tabular}{l|ccc}
        \hline
        \textbf{Method} & \textbf{Recall(\%)} & \textbf{Precision(\%)} & \textbf{F1(\%)} \\ 
        \hline
       Concatenate & \underline{97.3 }& 84.2 & 90.3 \\
       Element-wise Sum& 96.6 & 87.3 & 91.7 \\
       Element-wise Product & 94.9 & 86.0 &\underline{90.3}  \\
        \hline
        Cross-attention & \textbf{97.5} & \textbf{87.7} & \textbf{92.3} \\
    \hline
    \end{tabular}
    \label{ablation:crossattention}
    % \vspace{-2mm}
\end{table}

%  \begin{figure*}[t]
%     \centering
%     \includegraphics[width=0.95\linewidth]{image/analysis.png} 
%     \caption{Visualization of the t-SNE embedding and feature distribution for the test samples.}
%     \label{analysis}
%     % \vspace{-5mm}
% \end{figure*}

\subsection{Performance Analysis} 
We conduct a comparative analysis between the proposed method and DSVDD in the feature space. Fig. \ref{fig:main} (a) and (b) illustrate the variance distribution of the test sample features along principal components obtained via Principal Component Analysis (PCA). M-SVDD effectively concentrates normal samples within the principal component, while anomalies exhibit broader variance across secondary and the remaining components, thereby enhancing their separability. In contrast, DSVDD exhibits considerable overlap between anomalies and normal samples along the principal component, resulting in reduced discriminability. Furthermore, Fig. \ref{fig:main} (c) present the covariance matrix of the features for both normal and anomalous data. The results demonstrate that M-SVDD is able to effectively capture inter-feature correlations in normal data, whereas anomaly samples fail to exhibit such relationships, providing a clear distinction for anomaly detection.

\begin{figure}[t]
    \centering
    \includegraphics[width=0.80\linewidth]{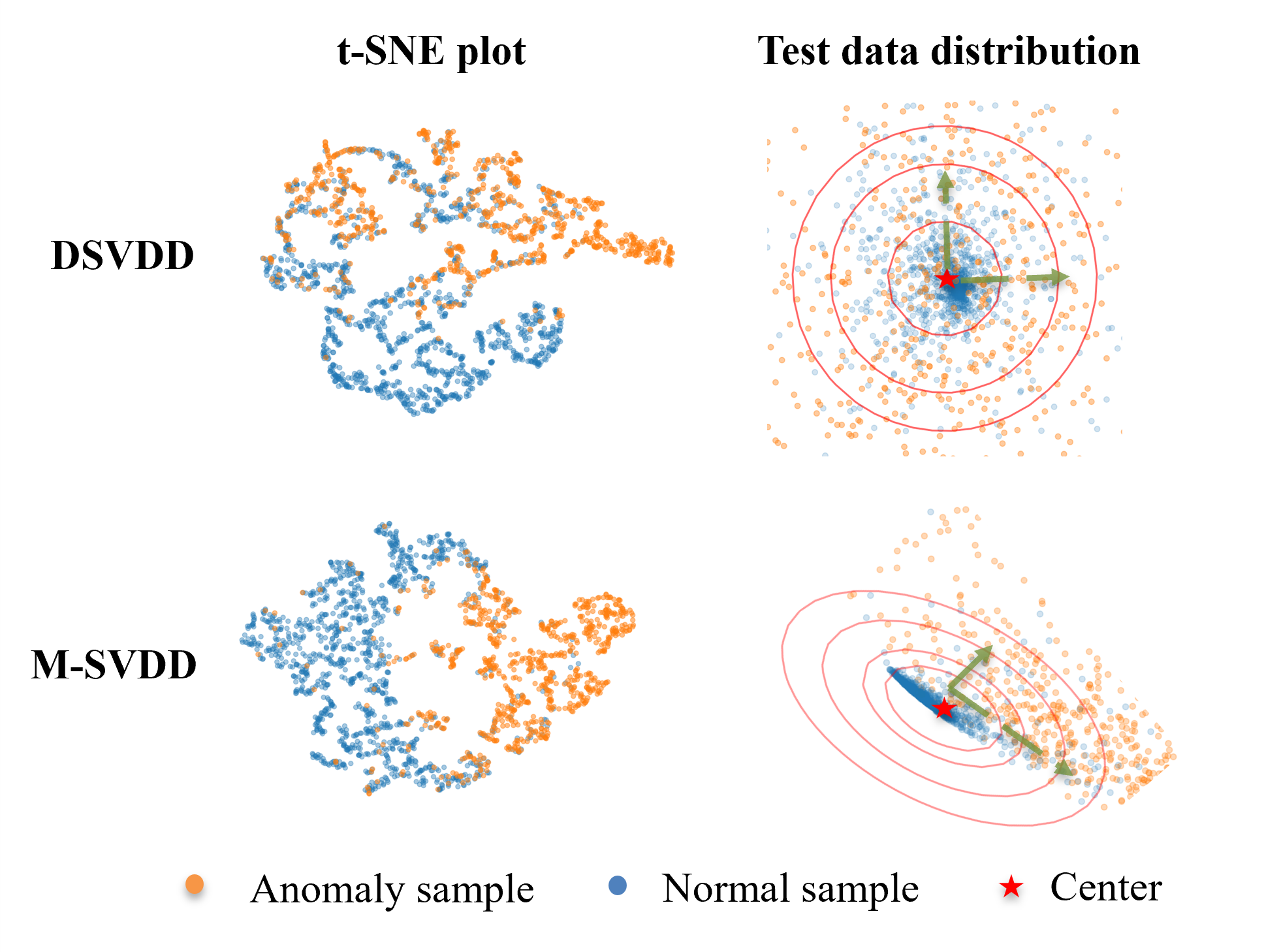}    
    \caption{Visualization of the t-SNE embedding and feature distribution for the test samples.}
    \label{feature_spcae}
    % \vspace{-3mm}
\end{figure}

To visualize the separability of the learned representations, we additionally train the network with the dimension of the hyperspace $s$ set to 2 and plot the feature distribution of the anomaly and normal data as shown in Fig. \ref{feature_spcae}. It shows that our method provides a more distinct separation between anomalies and normal data. In addition, t-distributed Stochastic Neighbor Embedding (t-SNE) projections show that features from the M-SVDD model form more compact and well-separated clusters, highlighting its superior ability to learn discriminative representations to identify anomalous data.

\begin{figure}[t]
    \centering
    \includegraphics[width=1.0\linewidth]{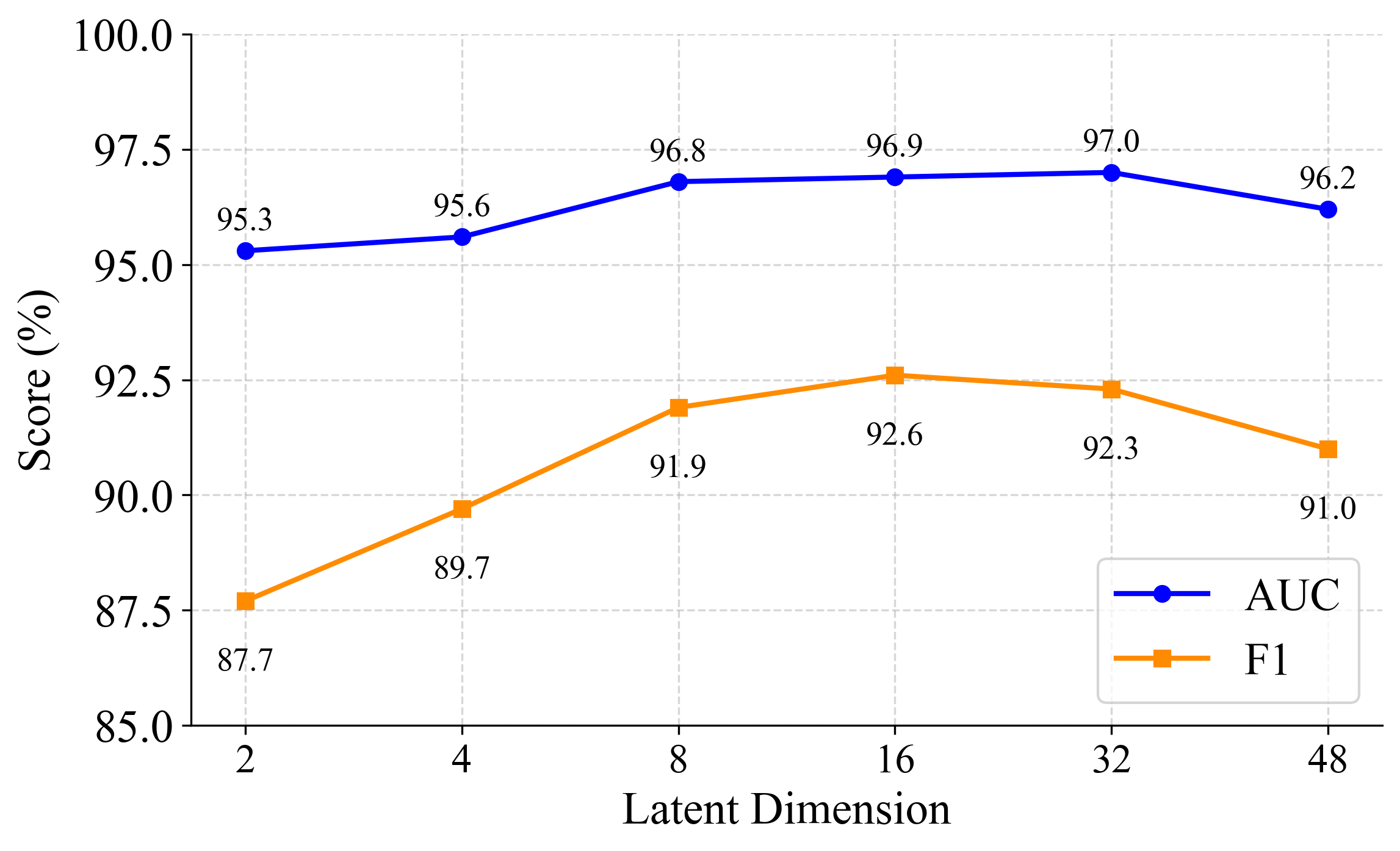}    \caption{Evaluation of the M-SVDD's performance across different latent space dimensions.}
    \label{latent analysis}
    % \vspace{-3mm}
\end{figure}

We further analyze the impact of the hyperspace dimension, which directly affects the network ability to represent normal data and estimate the $\mu_z$ and $\Sigma_z$. The results presented in Fig.~\ref{latent analysis} show that increasing the latent dimension from 2 to 32 consistently improves performance, with both AUC and F1-score increasing significantly. This suggests that a higher-dimensional latent space enables more expressive representations and tighter elliptical boundaries around normal samples. However, further increasing $s$ beyond 32 leads to a slight performance drop, likely due to feature redundancy and increased difficulty in estimating a stable covariance matrix. Overall, setting $s$ between 16 and 32 offers a good trade-off between representation capacity and robustness.

\begin{figure*}[t]
    \centering
    \includegraphics[width=0.9\linewidth]{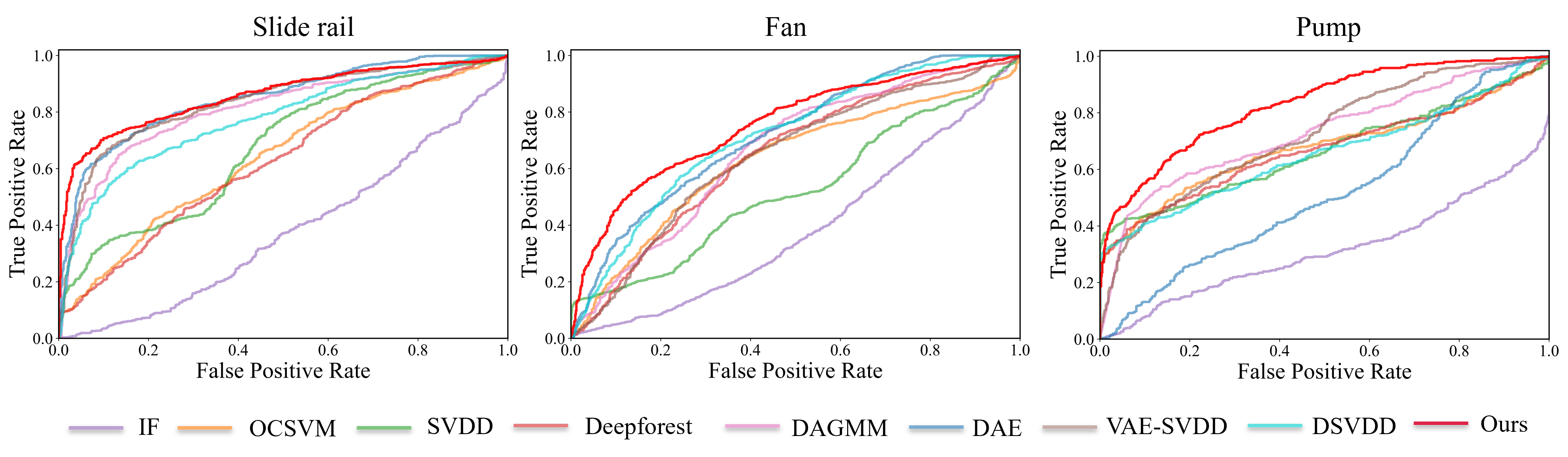}
    \caption{Comparison of ROC curves across different machine types on the MIMII dataset.}
    \label{auc_mimii}
    \vspace{-2mm}
\end{figure*}

\begin{table}[t]
\centering
\renewcommand{\arraystretch}{1.3}
\setlength{\tabcolsep}{9pt}  % Adjust column spacing
\caption{The AUC scores (\%) for different comparison methods on MIMII dataset.}
\begin{tabular}{l|cccc}
\hline
\textbf{Methods} & \textbf{Slide rail} & \textbf{Fan} & \textbf{Pump} &  \textbf{Average}\\
\hline
IF \cite{liu2008isolation} & 37.9	&39.0		&31.7	&36.2\\
OCSVM \cite{manevitz2001one} & 64.0	&62.9		&67.6	&64.8\\
SVDD \cite{tax2004support} & 67.4	 &52.4	  &66.6	 &62.1
 \\
DeepForest \cite{xu2023deep} & 62.5	&64.7		&66.9	&64.7
  \\
DAGMM \cite{zong2018deep}  & 81.1	&66.5		&\underline{72.8}	&73.5\\
DAE & \underline{84.9}  &  \underline{71.8} & 51.6 &69.4\\
VAE-SVDD  \cite{zhou2021vae} & {83.6}	&64.6		&72.5	&\underline{73.6}
  \\ 
DSVDD  \cite{ruff2018deep} & 77.6	&{71.6}&65.6	&71.6
  \\
\hline
\textbf{Ours} & \textbf{85.8}	&\textbf{75.6}	&\textbf{83.1}	&\textbf{81.5}
 \\
\hline
\end{tabular}
\label{comparison_results_mimii}
\end{table}
% \begin{figure}[t]
%     \centering
%     \includegraphics[width=0.8\linewidth]{image/data_samples_mimii.png}
%     \caption{Examples of selected samples in MIMII dataset.}
%     \label{data_samples_mimii}
%     % \vspace{-3mm}
% \end{figure}

\subsection{{Experiments on Public Datasets}} 
\textbf{Experiment on MIMII dataset\cite{purohit2019mimii}}: To further validate the generalization capability of the proposed M-SVDD-based unsupervised anomaly detection framework, we evaluate its performance on the MIMII dataset, which contains sound recordings from multiple machine types operating under both normal and faulty conditions. 
Specifically, we select the data collected from slide rail, fan and pump under 0dB noise setting to conduct the experiments. All audio signals are converted into spectrograms using a 1024-point FFT with a hop length of 512. We employ a two-dimensional CNN-based autoencoder for feature extraction and spectrogram reconstruction. Each comparison method is trained separately on each machine type, and the corresponding AUC scores are calculated for performance evaluation. 

The results in Table~\ref{comparison_results_mimii} show that the proposed M-SVDD framework achieves the best performance across all machine types. Specifically, our method obtains AUC scores of 85.8\%, 75.6\%, and 83.1\% on slide rail, fan, and pump, which are 0.9\%, 3.8\%, and 10.3\% higher than those of the second-best method.
On average, our method outperforms DSVDD and VAE-SVDD by 10.1\% and 7.9\% in AUC, respectively. In addition, the ROC curves in Fig.~\ref{auc_mimii} clearly show that our method achieves a better balance between the true positive rate and false positive rate, leading to more accurate and reliable anomaly detection across all machine types.

\textbf{Experiment on multi-dimensional time-series dataset}: The proposed unsupervised anomaly detection model is also evaluated on three multi-dimensional time-series anomaly dataset: Secure Water Treatment (SWaT)\cite{mathur2016swat}, Soil Moisture Active Passive (SMAP)\cite{hundman2018detecting} and Mars Science Laboratory (MSL) \cite{hundman2018detecting} to further verify the effectiveness of the proposed method. We use temporal convolutional network \cite{bai2018empirical} for feature extraction and adopt F1-score  to evaluate the performance of the networks. The proposed M-SVDD method is compared with TranAD\footnote{\url{https://github.com/xuhongzuo/DeepOD}}\cite{tuli2022tranad} and COUTA\footnotemark[5]\cite{xu2024calibrated}, which are specifically designed for timeseries data anomaly detection. We follow the data preprocessing and evaluation protocols used in TranAD\footnote{\url{https://github.com/imperial-qore/TranAD}}\cite{tuli2022tranad}, including the point-adjust evaluation strategy, which is also adopted in related works~\cite{xu2024calibrated,audibert2020usad}.

The experimental results are presented in Table \ref{comparison_results_pulic}. The proposed method achieves the highest F1-score on the SWaT and SMAP datasets, with figures of 82.0\% and 94.8\%, respectively, outperforming other comparison methods. On the MSL dataset, it achieves an F1-score of 98.1\%, which is only 0.5\% lower than the best-performing baseline (COUTA with 98.6\%), while still outperforming other deep learning approaches such as DSVDD and DeepForest. These results further demonstrate the robustness and generalization capability of our method across a variety of anomaly detection tasks.

\begin{table}[t]
\centering
\renewcommand{\arraystretch}{1.3}
\setlength{\tabcolsep}{9pt}  % Adjust column spacing
\caption{Comparison of methods based on F1-score (\%) across three time-series datasets.}
\begin{tabular}{l|ccc}

\hline
\textbf{Methods} & \textbf{SWaT}\cite{mathur2016swat} & \textbf{SMAP}\cite{hundman2018detecting} & \textbf{MSL}\cite{hundman2018detecting} \\
\hline
IF \cite{liu2008isolation}          & 29.6      & 89.3 &   92.6    \\
OCSVM \cite{kriegel2008angle}       & 24.2       & 91.1    & 93.9         \\
SVDD \cite{tax2004support}        & 80.7     &   94.1  &97.9 \\ 
DeepForest \cite{xu2023deep}       & 77.8          & 81.4  &  78.4       \\
TranAd \cite{tuli2022tranad}       & 81.4        & 91.0    & 95.1 \\
COUTA \cite{xu2024calibrated}      & \underline{81.4}       & \underline{93.8}    & \textbf{98.6} \\
DSVDD \cite{ruff2018deep}         & 81.5        & 92.5   & 98.0 \\
\hline
\textbf{Ours}                      & \textbf{82.0}         & \textbf{94.8}    & \underline{98.1} \\
\hline
\end{tabular}
\label{comparison_results_pulic}

\end{table}

\section{Limitation and Future work}
% While the proposed method demonstrates strong performance in detecting anomalies in mobile robots, false positives may still occur when both modalities are affected by significant noise. For instance, when the robot drives over uneven terrain or small obstacles, audio and IMU signals may deviate from the normal distribution, leading to incorrect anomaly alerts. False negatives may also occur when collisions or mechanical faults produce weak or imperceptible signals, leading the network to classify anomalous inputs as normal and potentially miss critical failures.

% To address these limitations, we plan to incorporate additional sensing modalities to enhance the system’s perception of the environment and internal states. Furthermore, identifying the specific type of anomaly is essential for enabling appropriate responses. As future work, we aim to integrate a large language model to support zero-shot anomaly classification, enabling more interpretable and actionable outputs.

While the proposed method demonstrates strong performance in detecting anomalies in mobile robots, several limitations remain due to the complexity of real-world sensor data. False positives may occur when both sensing modalities are simultaneously affected by external noise. For example, when the robot moves over uneven terrain or small obstacles, both IMU and audio signals may deviate from their normal patterns and trigger incorrect alerts. Another limitation is the inconsistency and timing misalignment between sensing modalities, which can result from sensor jitter, buffer overflow, or unsynchronized data collection. This issue is particularly common in systems built on the ROS2 framework \cite{li2025robospike}, where different nodes may assign timestamps using independent clocks or inconsistent message headers. These imperfections introduce temporal misalignment between inputs, leading to distributional shifts that reduce detection accuracy. Additionally, the current system does not incorporate contextual reasoning, which limits its ability to differentiate between harmless disturbances such as surface vibration and actual hardware faults.

To overcome these challenges, future work will focus on improving both robustness and interpretability. One approach is to incorporate cross-modal alignment methods that can compensate for timing mismatches and better synchronize the input streams. Another direction is to design models that can reason about the context of an event by combining temporal and semantic features, allowing the system to more accurately judge the cause and severity of an anomaly. Furthermore, introducing additional sensing modalities such as visual or thermal data may help the system detect subtle or environment-specific issues that audio and IMU alone cannot capture. Finally, we plan to integrate a large language model to support zero-shot anomaly classification. This would allow the system to produce more interpretable results and to identify anomaly types even without extensive labeled training data, making the outputs more actionable in real-world robotic applications.

\section{Conclusion}
This paper proposes a novel unsupervised anomaly detection framework for autonomous mobile robots based on audio and IMU data. Without using any labeled anomaly data for training, the network is able to effectively detect system anomalies such as collisions and internal mechanical faults. Specifically, the framework employs a M-SVDD module that encapsulates normal features within an ellipsoidal space and capture inter-feature correlations, enabling more precise and robust anomaly detection. In addition, a reconstruction branch is incorporated to mitigate the hypersphere collapse issue and improve detection accuracy.
Experimental evaluations conducted on a collected mobile robot dataset and four public datasets verify the effectiveness of the proposed approach.

% \begin{figure}[t]
%     \centering
%     \includegraphics[width=1.0\linewidth]{image/mimi_chart.png}
%     \caption{The inference results of test samples in the dataset.}
%     \label{mimi_chart}
%     % \vspace{-3mm}
% \end{figure}

\section*{Acknowledgments}
This research is supported by the National Research Foundation, Singapore, under its Medium-Sized Center for Advanced Robotics Technology Innovation (CARTIN).


\bibliographystyle{./IEEEtran}
\bibliography{./IEEEabrv,./AI_AD}

\vfill

\end{document}